\documentclass{article}
\usepackage{iclr2026_conference,times}

\usepackage{amsmath,amsfonts,bm}

\def\eqref#1{equation~\ref{#1}}

\def\1{\bm{1}}

\DeclareMathAlphabet{\mathsfit}{\encodingdefault}{\sfdefault}{m}{sl}
\SetMathAlphabet{\mathsfit}{bold}{\encodingdefault}{\sfdefault}{bx}{n}

\usepackage{array}
\usepackage{booktabs}
\usepackage{colortbl}
\usepackage{graphicx}
\usepackage{float}
\usepackage{longtable}
\usepackage{multirow}
\usepackage{wrapfig}
\usepackage{soul}
\usepackage{subcaption}
\usepackage{url}
\usepackage{hyperref}

\definecolor{FindingShade}{gray}{0.90}
\sethlcolor{FindingShade}
\newcommand{\findingheadline}[1]{%
    \par\smallskip
    \noindent{\bfseries\boldmath\hl{#1}}%
    \hspace{0.45em}\ignorespaces
}

\title{LLaDA MoE v2: Scaling Mixture-of-Experts Diffusion Language Models}

\author{
  Fengqi Zhu$^{1, 2, 3, \mathparagraph}$, Shaoxuan Xu$^{1, 2, 3}$, Jingyang Ou$^{1, 2, 3}$, Zebin You$^{1, 2, 3}$, Yipeng Xing$^{4}$, \\
  \textbf{Huabin Liu}$^{4}$\textbf{,} \textbf{Xiaolu Zhang}$^{4}$\textbf{,} \textbf{Jun Zhou}$^{4}$\textbf{,} \textbf{Zhenzhong Lan}$^{4}$\textbf{,} \textbf{Yankai Lin}$^{1, 2, 3}$\textbf{,} \\
  \textbf{Wayne Xin Zhao}$^{1, 2, 3}$\textbf{,} \textbf{Jianguo Li}$^{4}$\textbf{,} \textbf{Chongxuan Li}$^{1, 2, 3, \ddagger}$\textbf{,} \textbf{Ji-Rong Wen}$^{1, 2, 3, \ddagger}$  \\
  $^1$ Gaoling School of AI, Renmin University of China \\
  $^2$ Beijing Key Laboratory of Research on Large Models and Intelligent Governance \\
  $^3$ Engineering Research Center of  Next-Generation Intelligent Search and Recommendation, MOE \\
  $^4$ Ant Group \\
}

\iclrfinalcopy
\begin{document}

\maketitle

\begin{abstract}
Diffusion language models (dLLMs) offer an alternative to autoregressive (AR) language modeling, yet the scaling behavior of Mixture-of-Experts (MoE) dLLMs remains poorly understood. We systematically characterize how optimization hyperparameters, compute allocation, and architecture scale for MoE dLLMs, identifying quantitative differences from scaling trends previously reported for AR models. Specifically, for optimization, the optimal nominal batch size grows faster, while the optimal learning rate decays more rapidly with compute. For model--data allocation, IsoFLOP analysis reveals a slight data-side tilt: the optimal token budget grows faster than activated model-side computation. For MoE architecture, larger scales increasingly favor larger expert pools at fixed activated capacity, while moderate expert granularity remains consistently effective and the preferred fraction of activated capacity assigned to shared experts remains stable across scales. Guided by these findings, we train LLaDA MoE v2, a 30B-A3B dLLM, from scratch on 23.5T tokens. With approximately 65\% as many pretraining tokens as Qwen3, LLaDA MoE v2 approaches Qwen3 on several knowledge, reasoning, and coding benchmarks. After supervised fine-tuning alone, it outperforms SDAR Chat on seven of eight reasoning and coding benchmarks and remains close to Qwen3 on several tasks. These results establish practical scaling laws and design principles for MoE dLLMs.
\end{abstract}

\makeatletter
\renewcommand{\@fnsymbol}[1]{\ensuremath{%
  \ifcase#1\or
    *\or
    \dagger\or
    \ddagger\or
    \mathsection\or
    \mathparagraph\or
    \|\or
    **\or
    \dagger\dagger\or
    \ddagger\ddagger
  \else
    \@ctrerr
  \fi}}
\makeatother

\renewcommand{\thefootnote}{\fnsymbol{footnote}}
\footnotetext[5]{Work done during an internship at Ant Group}
\footnotetext[3]{Corresponding author}
\renewcommand{\thefootnote}{\arabic{footnote}}

\section{Introduction}

Large language models~\citep{zhao2026survey} have advanced largely through scale: as models, data, and compute grow, performance improves in systematic and predictable ways~\citep{kaplan2020scaling,hoffmann2022training}. Most of this progress has followed the autoregressive (AR) paradigm~\citep{radford2018improving,radford2019language,brown2020language,ouyang2022training}, which factorizes the distribution of a text sequence into next-token conditionals along a fixed left-to-right order. Diffusion language models (dLLMs)~\citep{lou2023discrete,sahoo2024simple,ou2025your,nie2026large} offer an alternative probabilistic formulation: they define the distribution through an iterative denoising process, in which a bidirectional model reconstructs the masked tokens of a corrupted sequence, allowing multiple tokens to be decoded in parallel at each step~\citep{wu2026fast,li2026refusion,chen2026dflash}. Recent dLLMs now match the capabilities of strong AR models, both when trained from scratch~\citep{nie2025scaling,nie2026large,nie2026improved} and when adapted from pretrained AR checkpoints~\citep{bie2025llada2,dream2025}, making them a promising candidate for future language modeling.

Realizing this potential, however, requires understanding how dLLMs behave as training budgets grow. Existing scaling studies of dLLMs have so far focused on dense architectures~\citep{nie2025scaling,ni2025training,von2026scaling,sahoo2026scaling}. The AR literature, in contrast, has widely adopted Mixture-of-Experts (MoE) Transformer architectures~\citep{vaswani2017attention,shazeer2017outrageously,lepikhin2020gshard,fedus2022switch}, which expand model capacity far beyond what dense models can afford at the same computation per token~\citep{du2022glam,dai2024deepseekmoe,liu2024deepseekv2,liu2024deepseekv3,tian2026towards,yang2025qwen3}. MoE dLLMs have therefore begun to emerge~\citep{zhu2025llada,feng2026dmoe,zhang2026expert}, but their designs largely inherit AR practice. AR experience provides a useful prior, but it cannot be assumed to transfer directly: dLLMs optimize a masked denoising objective rather than next-token prediction, supervision falls only on masked positions, and each prediction conditions on a corrupted sequence rather than a causal prefix. Consequently, the scaling behavior of MoE dLLMs remains undercharacterized.

In this work, we systematically characterize the scaling behavior and architectural design of MoE dLLMs across compute scales. We ask how the optimal batch size and learning rate change with compute, how a fixed compute budget should be allocated between activated model-side computation and training tokens, and how the resulting activated budget should be decomposed into routing sparsity, expert granularity, and shared capacity. Table~\ref{tab:findings-overview} summarizes the resulting empirical findings and the corresponding evidence. Across these dimensions, AR experience provides a useful prior but requires dLLM-specific calibration.

\begin{table}[t]
    \centering
    \small
    \setlength{\tabcolsep}{3pt}
    \renewcommand{\arraystretch}{1.10}
    \caption{\textbf{Empirical findings for MoE dLLMs.} Controlled sweeps cover optimization, compute allocation, and MoE architecture, with scale-up evidence from LLaDA MoE v2.}
    \label{tab:findings-overview}
    \begin{tabular}{@{}>{\raggedright\arraybackslash}p{0.16\linewidth}>{\raggedright\arraybackslash}p{0.68\linewidth}>{\raggedright\arraybackslash}p{0.10\linewidth}@{}}
        \toprule
        \textbf{Category} & \textbf{Finding} & \textbf{Evidence} \\
        \midrule
        \multirow{2}{*}{Optimization} & Optimal batch size grows more steeply with compute than in AR models. & Figs.~\ref{fig:hyperparameter-scaling},~\ref{fig:batchsize-lr-grid} \\
                                      & Optimal learning rate decays faster with compute than in AR models. & Figs.~\ref{fig:hyperparameter-scaling},~\ref{fig:batchsize-lr-grid} \\
        \addlinespace[8pt]
        \multirow{2}{*}{\shortstack[l]{Compute\\allocation}} & Optimal compute allocation is near-balanced with a slight data-side tilt. & Fig.~\ref{fig:compute-allocation} \\
                                                              & The MoE dLLM frontier is more data-tilted than dense dLLM frontiers. & Table~\ref{tab:allocation-scaling-laws} \\
        \addlinespace[8pt]
        \multirow{3}{*}{\shortstack[l]{MoE\\architecture}} & Lower activation ratios are increasingly favorable at larger scales. & Fig.~\ref{fig:architecture-activation} \\
                                                           & Moderate expert granularity, \(G=8\)--\(16\), is robust across scales. & Fig.~\ref{fig:architecture-granularity} \\
                                                           & The optimal shared-expert ratio stays at \(S=33.3\%\) across scales. & Fig.~\ref{fig:architecture-shared} \\
        \addlinespace[8pt]
        \multirow{3}{*}{\shortstack[l]{Scale-up\\validation}} & Approaches Qwen3 on some tasks with 35\% fewer pretraining tokens. & Table~\ref{tab:pretrain-results} \\
                                                              & Matches or exceeds 7B-A1B at substantially lower compute. & Fig.~\ref{fig:benchmark_scaling_comparison} \\
                                                              & Outperforms SDAR Chat on seven of eight tasks after SFT. & Table~\ref{tab:sft-results} \\
        \bottomrule
    \end{tabular}
\end{table}

These findings yield practical guidance for scaling MoE dLLMs, which we validate by training LLaDA MoE v2, a 30B-A3B model, from scratch on 23.5T tokens. LLaDA MoE v2 approaches Qwen3 30B-A3B~\citep{yang2025qwen3} on several knowledge, reasoning, and coding benchmarks while using approximately 65\% as many pretraining tokens, and matches a prior MoE dLLM~\citep{zhu2025llada} developed without such guidance at a fraction of its training compute. Standard SFT turns it into a strong instruct model: LLaDA MoE v2 outperforms SDAR Chat 30B-A3B~\citep{cheng2026sdar} on seven of eight reasoning and coding benchmarks and remains competitive with Qwen3 across multiple benchmarks despite Qwen3's additional RL stage.

\section{Preliminaries}
\label{sec:preliminaries}

\subsection{Diffusion Language Models}

Masked diffusion language models~\citep{austin2021structured,campbell2022continuous,lou2023discrete,shi2024simplified} define a discrete diffusion process over token sequences. Let $x_0 = (x_0^1, \ldots, x_0^L)$ denote a clean sequence of $L$ tokens, and let \texttt{[MASK]} be the mask token. The forward process corrupts each position independently: at noise level $t \in [0, 1]$, the token $x_0^i$ is replaced by \texttt{[MASK]} with probability $t$ and kept unchanged otherwise, producing the corrupted state $x_t$.

A bidirectional Transformer~\citep{vaswani2017attention} $p_\theta$ is trained to reverse this process, predicting the clean token at each masked position conditioned on the corrupted state. With the noise level sampled uniformly, training minimizes the denoising objective
\begin{equation}
    \mathcal{L}(\theta) = -\mathbb{E}_{x_0,\, t,\, x_t} \left[ \frac{1}{t} \sum_{i=1}^{L} \mathbf{1} \left[x_t^i = \text{[MASK]}\right] \log p_\theta\left(x_0^i \mid x_t\right) \right],
    \label{eq:dllm-objective}
\end{equation}
which upper-bounds the negative log-likelihood of the data. At inference time, generation starts from a fully masked sequence and unmasks tokens through iterative denoising steps.

\subsection{Mixture-of-Experts Transformers and Scaling Laws}

A Mixture-of-Experts (MoE) Transformer decouples model capacity from per-token computation by replacing the feed-forward network with $n_e$ routed experts and a lightweight router~\citep{jacobs1991adaptive,shazeer2017outrageously,lepikhin2020gshard,fedus2022switch,zoph2022st,dai2024deepseekmoe}: the total parameter count scales with $n_e$, whereas each token is processed by only a small subset of them. Concretely, the router activates the top-$n_a$ routed experts for each token and combines their outputs. In addition, every token is processed by a single shared expert with intermediate width $d_{\mathrm{share}}=n_s d_{\mathrm{expert}}$, where $n_s$ denotes shared capacity measured in units of one routed expert. We parameterize such an architecture by three variables: the activation ratio $A = (n_a + n_s)/(n_e + n_s)$, which is the fraction of expert capacity activated per token; the expert granularity $G = 2d_{\mathrm{model}}/d_{\mathrm{expert}}$, which controls how routed capacity is partitioned into experts; and the shared-expert ratio $S = n_s/(n_a + n_s)$, which controls the fraction of activated expert capacity assigned to the shared pathway~\citep{krajewski2024scaling,abnar2025parameters,tian2026towards}.

Scaling laws use small-scale measurements to estimate how compute-dependent choices, including optimization hyperparameters, model--data allocation, and architecture, should change at larger budgets~\citep{hestness2017deep,kaplan2020scaling,hoffmann2022training,krajewski2024scaling,abnar2025parameters}. In practice, power-law fits model the optimal batch size and learning rate as functions of the compute budget \(C\), while IsoFLOP analysis fixes \(C\) and sweeps model-side computation against training tokens~\citep{hoffmann2022training,bi2024deepseek,li2025predictable}. Dense-model studies commonly approximate training compute as \(C\approx6ND\), where \(N\) is the number of non-embedding parameters and \(D\) is the number of training tokens, assuming that all parameters participate in processing each token~\citep{kaplan2020scaling,hoffmann2022training}. Because this assumption does not hold for MoE models, we follow prior MoE scaling studies and use \(C=MD\), where \(M\) denotes activated non-embedding training FLOPs per token~\citep{bi2024deepseek,ludziejewski2025joint,tian2026towards}; full compute expressions are provided in Appendix~\ref{app:compute-accounting}.

\section{Scaling Laws for MoE dLLMs}

In this section, we develop a scaling framework for MoE dLLMs in three stages. We first calibrate compute-dependent batch size and learning rate, then estimate the optimal allocation between activated model-side computation and training data, and finally decompose the resulting model-side budget into activation ratio, expert granularity, and shared-expert ratio.

\subsection{Scaling Law for Hyperparameters}

\begin{figure}[t]
    \centering
    \begin{minipage}[t]{0.49\linewidth}
        \centering
        \includegraphics[width=\linewidth, trim=0 5pt 0 0, clip]{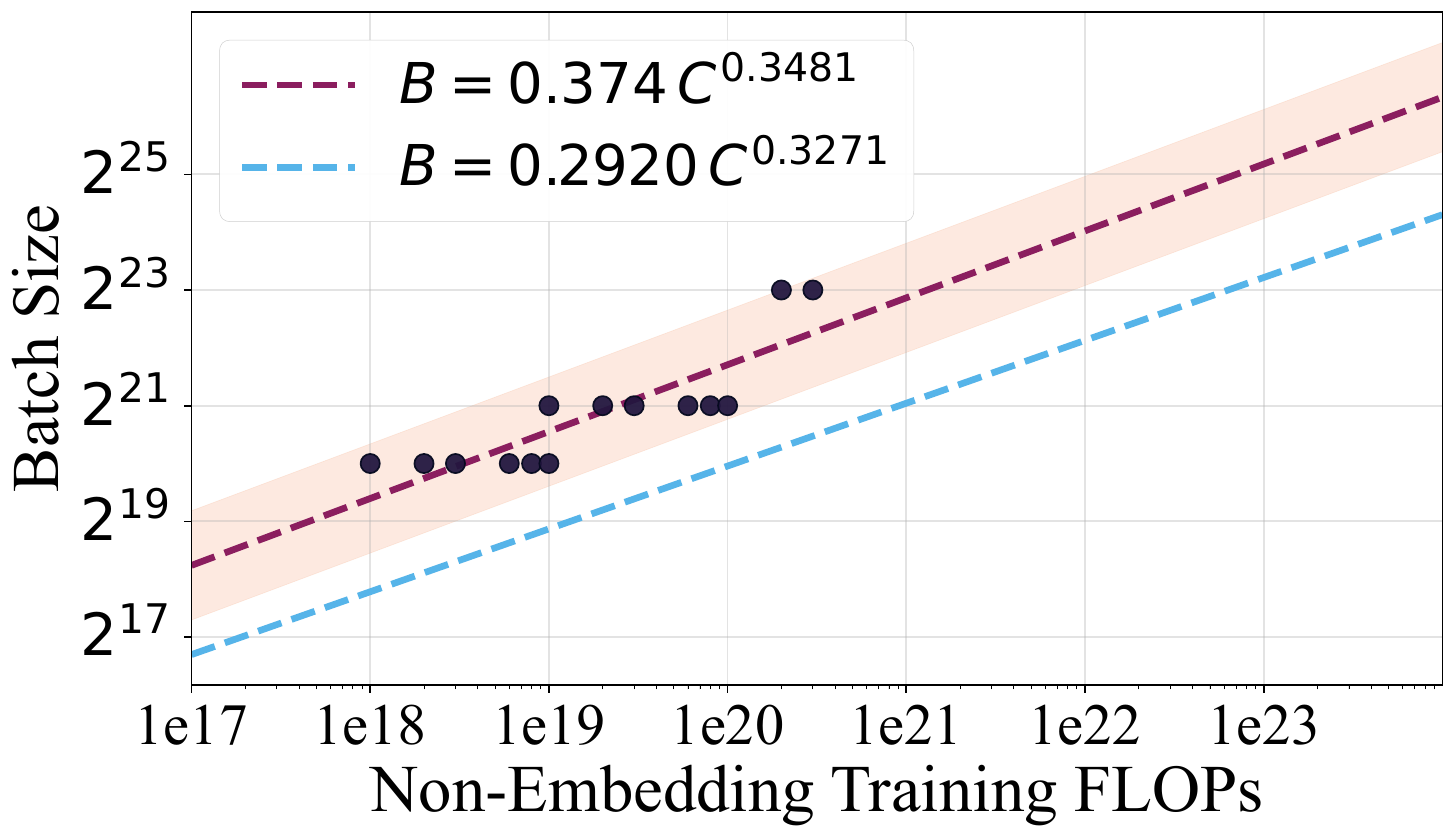}
    \end{minipage}
    \hfill
    \begin{minipage}[t]{0.49\linewidth}
        \centering
        \includegraphics[width=\linewidth, trim=0 5pt 0 0, clip]{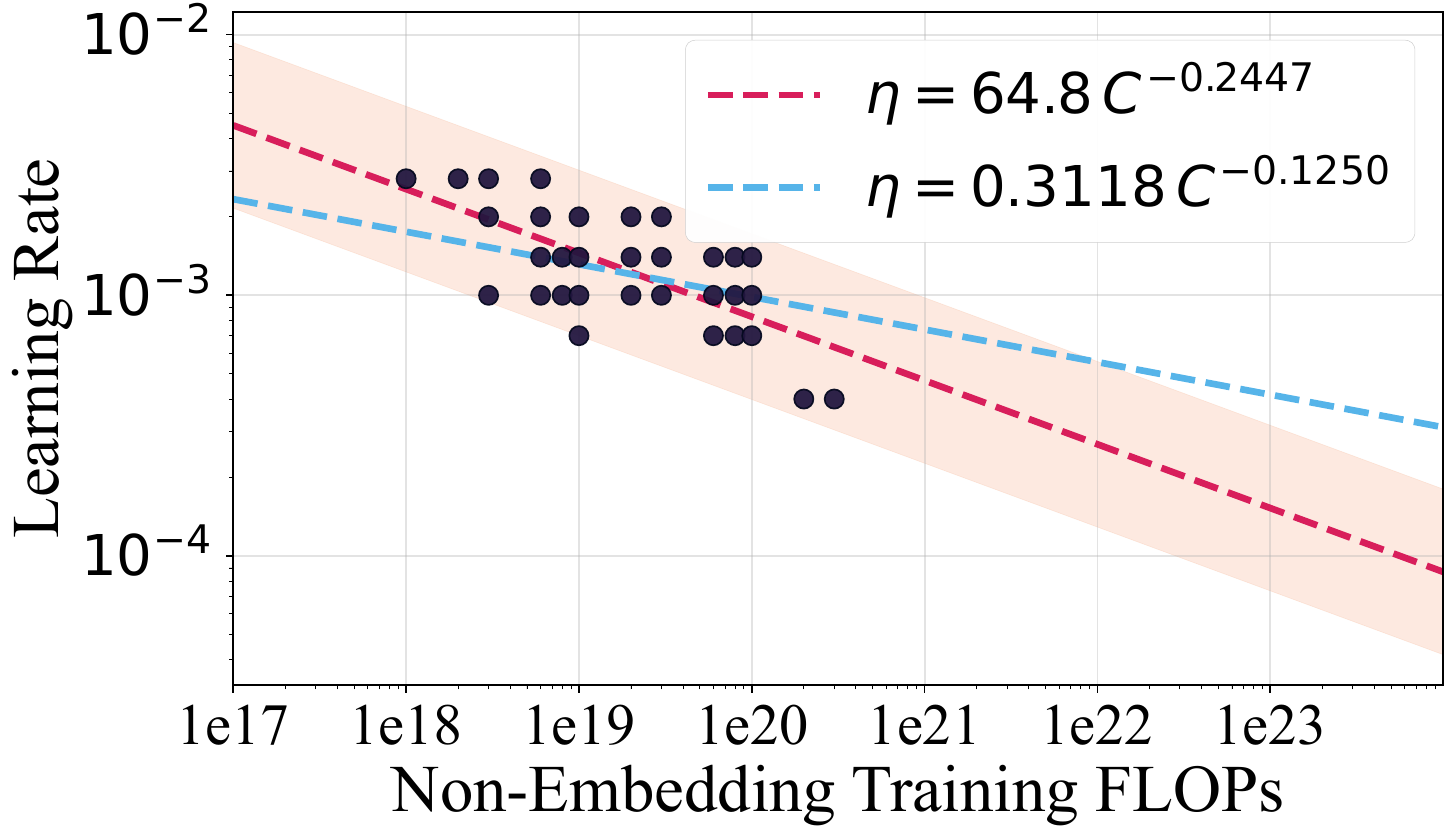}
    \end{minipage}
    \caption{\textbf{Scaling curves of nominal token batch size and learning rate with training compute.} \textbf{Left:} nominal token batch size. \textbf{Right:} learning rate. Magenta dashed lines indicate our fitted scaling laws, blue dashed lines show the reference scaling laws from DeepSeek LLM~\citep{bi2024deepseek}, and shaded regions denote the empirical ranges around our fitted curves.}
    \label{fig:hyperparameter-scaling}
\end{figure}

Although hyperparameter scaling has been extensively studied for AR models~\citep{bi2024deepseek,li2025predictable}, whether the same compute-dependent laws describe dLLMs remains an open question. Unlike AR training, which optimizes next-token prediction over causal prefixes, dLLMs are trained with a masked denoising objective over partially observed sequences. Since each update supervises only sampled masked positions, the nominal token batch size does not directly correspond to the effective number of prediction targets; for example, under common uniform timestep sampling, only half of the tokens are predicted in expectation. This reduced effective supervision can change the gradient noise level, optimization stability, and learning-rate sensitivity. We therefore study how the optimal batch size and learning rate vary with compute, both to compare their scaling behavior with AR expectations and to establish stable, compute-efficient settings for the subsequent analyses.

Concretely, we perform a hyperparameter search over nominal token batch size and learning rate across a set of representative model scales, ranging from 158M to 3.6B, under compute budgets from \(10^{18}\) to \(3\times10^{20}\); the detailed training and model settings are provided in Appendix~\ref{app:hyperparameter-scaling}.

As shown in Figure~\ref{fig:hyperparameter-scaling}, we fit the scaling curves of batch size and learning rate against training compute. The resulting scaling laws can be summarized as
\begin{equation}
B^* = 0.374 \cdot C^{0.3481}, \quad
\eta^* = 64.8 \cdot C^{-0.2447},
\end{equation}
where $C$ denotes the compute budget, and $B^*$ and $\eta^*$ denote the optimal nominal token batch size and learning rate under the dLLM training objective, respectively.

\begin{wrapfigure}[19]{r}{0.44\linewidth}
    \vspace{-0.6\baselineskip}
    \centering
    \includegraphics[width=\linewidth, trim=0 5pt 0 0, clip]{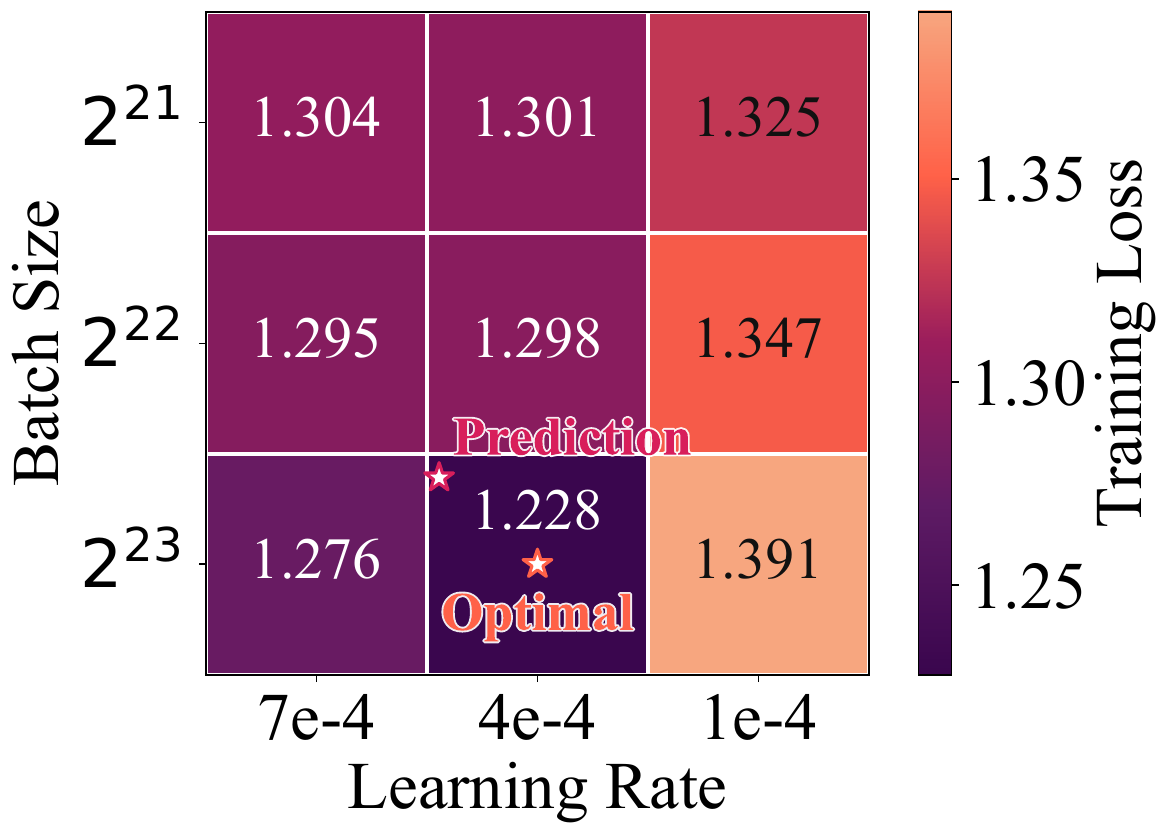}
    \caption{\textbf{Joint search over batch size and learning rate at \(6\times10^{20}\) FLOPs.}
    Each cell corresponds to one training run, with color and overlaid value denoting training loss; red stars mark the fitted scaling-law prediction and the best observed configuration.}
    \label{fig:batchsize-lr-grid}
    \vspace{-0.6\baselineskip}
\end{wrapfigure}

To test whether the fitted dLLM-specific scaling law extrapolates beyond its fitting range, we continue the joint batch-size and learning-rate runs from \(3\times10^{20}\) to \(6\times10^{20}\) FLOPs and examine the loss grid in Figure~\ref{fig:batchsize-lr-grid}. The fitted recommendation lies close to the best observed configuration in the two-dimensional hyperparameter plane, and the surrounding losses do not indicate a substantially better alternative within the searched settings. This result supports using the fitted scaling law as a practical estimate of the compute-dependent hyperparameters at this scale.

\findingheadline{Optimization: high compute favors larger batches and faster learning-rate decay.} The optimization directions are consistent between dLLM and AR training, with the optimal batch size growing sublinearly with training compute and the optimal learning rate decreasing at larger compute. The calibration, however, differs in a systematic way. Compared with DeepSeek LLM's AR law~\citep{bi2024deepseek}, our dLLM fit uses a slightly steeper batch-size scaling and a faster learning-rate decay, shifting the high-compute estimate toward larger nominal batches and smaller learning rates. At a compute budget of \(10^{20}\) FLOPs, for example, DeepSeek LLM's AR law predicts a similar learning rate, \(9.86\times10^{-4}\) compared with \(8.27\times10^{-4}\) from our fit, but a much smaller optimal batch size, 1.02M tokens compared with 3.43M tokens. This batch-size gap is consistent with the reduced effective supervision per nominal token noted above. Thus, AR scaling can provide a useful prior, but it should not replace dLLM-specific calibration when the fitted trends diverge. All subsequent experiments therefore use our fitted dLLM hyperparameter laws.

\subsection{Scaling Law for Compute Allocation}

Once the optimization hyperparameters are calibrated, the next issue is how to split a fixed compute budget between activated model-side computation and training data. For MoE dLLMs, this trade-off differs from its AR counterpart: each nominal token contributes a prediction target only when masked, while the router acts on corrupted states that vary with noise level and mask pattern rather than causal prefixes. Relative to AR training, a fixed nominal token budget therefore supplies fewer supervised targets over a more variable set of conditioning and routing states. Additional tokens can improve coverage of both denoising targets and router inputs, giving data-side investment a dLLM-specific marginal value; AR allocation laws thus cannot be assumed to transfer directly. Prior IsoFLOP studies characterize this trade-off for dense and MoE AR models~\citep{hoffmann2022training,bi2024deepseek,ludziejewski2025joint}, whereas analogous studies of dLLMs have mainly considered dense architectures, leaving the allocation behavior of MoE dLLMs uncharacterized.

To address this gap, we measure the model side by activated non-embedding FLOPs per token \(M\), which directly captures activated model computation; the data side is the number of training tokens \(D\), giving \(C=MD\). We sweep their allocation across fixed compute budgets ranging from \(10^{17}\) to \(10^{20}\) training FLOPs, with detailed settings provided in Appendix~\ref{app:compute-allocation}. For each budget, we identify the lowest-loss allocation within the sweep and fit the resulting frontiers \(M^*(C)\) and \(D^*(C)\).

\begin{table}[t]
    \centering
    \small
    \setlength{\tabcolsep}{6pt}
    \caption{\textbf{Representative compute-allocation scaling laws for language models.} We position our MoE dLLM frontier against representative model--data allocation laws across modeling objectives and architectures. The frontier columns report fitted growth exponents. For DLMs, we report the masked setting matching our formulation; the SMDM coefficients are taken from the DLMs analysis.}
    \label{tab:allocation-scaling-laws}
    \begin{tabular*}{\linewidth}{@{\extracolsep{\fill}}lllll@{}}
        \toprule
        Scaling law & Modeling & Architecture & Model frontier & Data frontier \\
        \midrule
        Kaplan~\citep{kaplan2020scaling} & AR & Dense & \(M^* \propto C^{0.73}\) & \(D^* \propto C^{0.27}\) \\
        Chinchilla~\citep{hoffmann2022training} & AR & Dense & \(M^* \propto C^{0.49}\) & \(D^* \propto C^{0.51}\) \\
        DeepSeek LLM~\citep{bi2024deepseek} & AR & Dense & \(M^* \propto C^{0.5243}\) & \(D^* \propto C^{0.4757}\) \\
        Llama 3~\citep{grattafiori2024llama} & AR & Dense & \(M^* \propto C^{0.463}\) & \(D^* \propto C^{0.537}\) \\
        SMDM~\citep{nie2025scaling} & AR & Dense & \(M^* \propto C^{0.644}\) & \(D^* \propto C^{0.356}\) \\
        Ling~\citep{tian2026towards} & AR & Dense & \(M^* \propto C^{0.5422}\) & \(D^* \propto C^{0.4578}\) \\
        Ling~\citep{tian2026towards} & AR & MoE & \(M^* \propto C^{0.5095}\) & \(D^* \propto C^{0.4905}\) \\
        \midrule
        SMDM~\citep{nie2025scaling} & Diffusion & Dense & \(M^* \propto C^{0.634}\) & \(D^* \propto C^{0.366}\) \\
        Quokka~\citep{ni2025training} & Diffusion & Dense & \(M^* \propto C^{0.514}\) & \(D^* \propto C^{0.486}\) \\
        DLMs~\citep{von2026scaling} & Diffusion & Dense & \(M^* \propto C^{0.566}\) & \(D^* \propto C^{0.434}\) \\
        Ours & Diffusion & MoE & \(M^* \propto C^{0.475}\) & \(D^* \propto C^{0.525}\) \\
        \bottomrule
    \end{tabular*}
\end{table}

\begin{figure}[t]
    \centering
    \includegraphics[width=\linewidth, trim=0 7pt 0 0, clip]{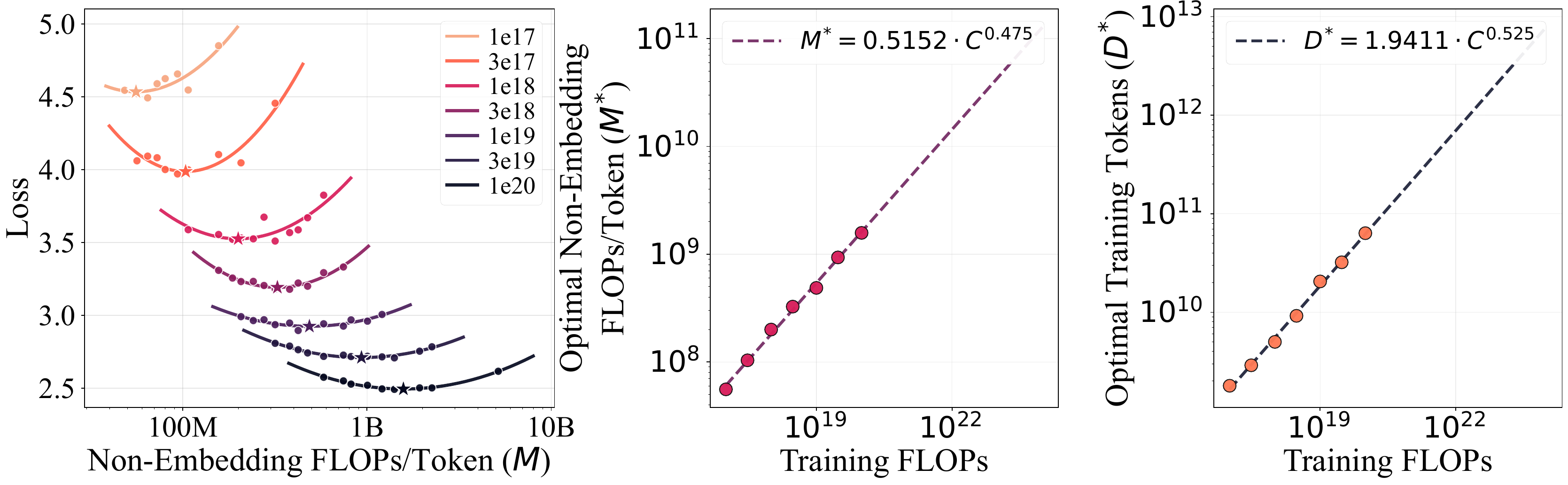}
    \caption{\textbf{IsoFLOP analysis for compute allocation.}
    \textbf{Left}: training loss under different model--data allocations at fixed compute budgets, with stars marking the lowest-loss. \textbf{Middle and right}: fitted power laws for the compute-optimal non-embedding FLOPs per token and training tokens.}
    \label{fig:compute-allocation}
\end{figure}

The left panel of Figure~\ref{fig:compute-allocation} shows the resulting U-shaped IsoFLOP curves across the compute budgets above: smaller models are capacity-limited despite seeing more tokens, while larger models become data-limited because fewer tokens can be trained under the same budget. The star-marked points give each budget's empirical compute-optimal allocation, \((M^*(C), D^*(C))\).

As the compute budget increases, these optima move toward larger activated non-embedding FLOPs per token and more training tokens, forming model-side and data-side allocation frontiers. The middle and right panels of Figure~\ref{fig:compute-allocation} plot \(M^*(C)\) and \(D^*(C)\), respectively; both follow approximately linear trends in log--log space. Fitting the two frontiers under the fixed-compute constraint gives
\begin{equation}
M^*(C)=0.5152\cdot C^{0.475},\quad D^*(C)=1.9411\cdot C^{0.525}.
\end{equation}
\findingheadline{Compute allocation: near-balanced scaling with a slight data-side tilt.} Both frontiers grow with compute, and their exponents are close to \(0.5\); the data-side exponent of \(0.525\) is slightly larger than the model-side exponent of \(0.475\). Thus, the optimal token budget grows faster than activated model-side computation. We contextualize this tilt against representative compute-allocation laws for AR models and dLLMs in Table~\ref{tab:allocation-scaling-laws}.

Within this comparison, two matched results help separate the effects of architecture and modeling objective. First, Ling compares dense and MoE architectures under AR modeling: moving from dense AR to MoE AR shifts the model/data exponents from \(0.5422/0.4578\) to \(0.5095/0.4905\), indicating that sparse activation moves the AR frontier toward more data. Second, SMDM compares AR and dLLM objectives under a dense architecture: moving from AR to dLLM shifts the exponents from \(0.644/0.356\) to \(0.634/0.366\), showing the same data-favoring direction. Our setting combines sparse activation with the dLLM objective, providing a lens for interpreting the data-side tilt of our frontier. While existing dense dLLM frontiers remain model-side dominated, our MoE dLLM frontier reaches a larger data-side exponent of \(0.525\). This result is consistent with the data-favoring tendencies associated separately with sparse activation and diffusion modeling in the comparisons.

The fitted frontier yields a simple allocation rule: the marginal compute of an MoE dLLM is best spent relatively more on additional training tokens than on increasing activated non-embedding FLOPs per token. Following this frontier fixes the aggregate activated model-side budget \(M^*(C)\) at each compute scale while leaving expert count, expert size, and activation pattern undetermined.

\subsection{Scaling Law for MoE Architecture}

Translating this fixed budget into an MoE architecture poses a dLLM-specific design problem because the router acts on masked denoising states that vary with noise level and mask pattern. We consider the three architectural dimensions defined in Section~\ref{sec:preliminaries}: the activation ratio \(A\), which controls routing sparsity and thereby the size of the routed expert pool; the expert granularity \(G\), which trades off routing diversity against per-expert capacity; and the shared-expert ratio \(S\), which determines the fraction of activated capacity assigned to a shared pathway. We examine how these dimensions should be configured across compute scales.

For each reference compute scale \(C\), we fix the activated model-side budget to \(M^*(C)\) and sweep one architectural dimension at a time while holding the other two fixed. Detailed settings are provided in Appendix~\ref{app:architecture-scaling}. Figure~\ref{fig:architecture} summarizes these controlled architecture sweeps.

\begin{figure}[t]
    \centering
    \includegraphics[width=\linewidth, trim=0 31pt 0 0, clip]{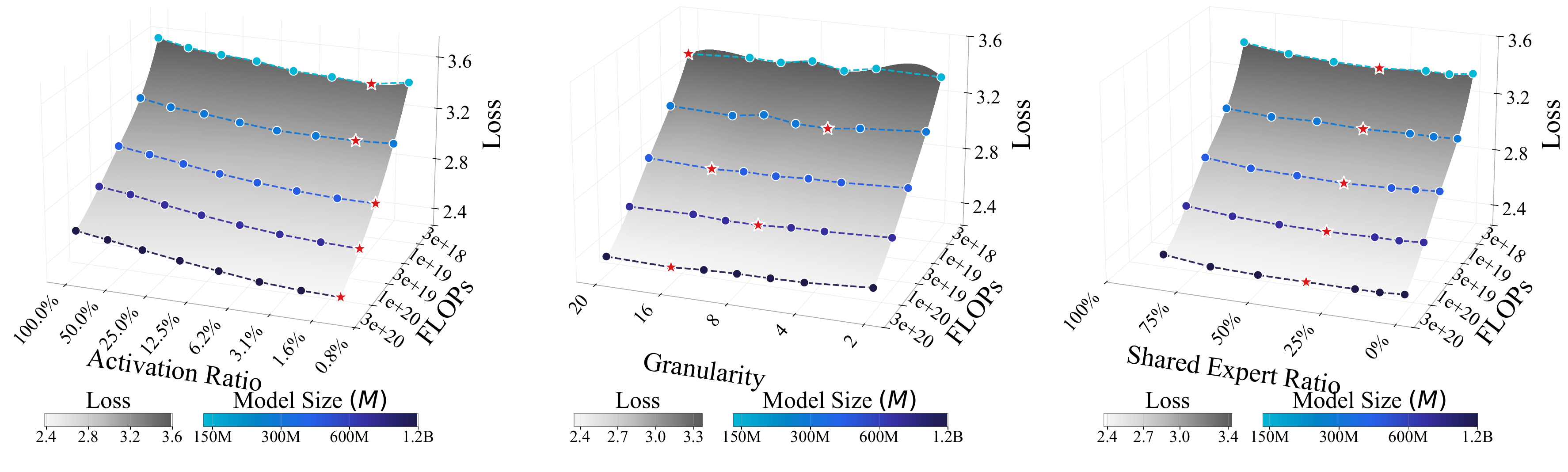}
    \vspace{-0.6em}
    \makebox[\linewidth][l]{%
        \hspace*{-0.012\linewidth}%
        \begin{minipage}[t]{0.32\linewidth}
            \phantomsubcaption\label{fig:architecture-activation}
            \centering\small\textbf{(a)} Activation Ratio (A)
        \end{minipage}%
        \hspace{0.02\linewidth}%
        \begin{minipage}[t]{0.32\linewidth}
            \phantomsubcaption\label{fig:architecture-granularity}
            \centering\small\textbf{(b)} Expert Granularity (G)
        \end{minipage}%
        \hspace{0.02\linewidth}%
        \begin{minipage}[t]{0.32\linewidth}
            \phantomsubcaption\label{fig:architecture-shared}
            \centering\small\textbf{(c)} Shared-Expert Ratio (S)
        \end{minipage}%
    }
    \caption{\textbf{MoE architecture scaling.}
    We evaluate how different decompositions of the activated model-side budget affect training loss across compute scales.
    \textbf{(a):} Activation Ratio \(A\), \textbf{(b):} Expert Granularity \(G\), and \textbf{(c):} Shared-Expert Ratio \(S\).
    Colors indicate the activated model-side budget \(M^*(C)\), and red stars mark the lowest-loss configurations at each compute scale.}
    \label{fig:architecture}
\end{figure}

\findingheadline{Activation ratio: larger scales favor sparser activation.}
We examine the activation ratio \(A\) in Figure~\ref{fig:architecture}(\subref{fig:architecture-activation}). Under the fixed activated model-side budget \(M^*(C)\), reducing \(A\) generally lowers training loss, with the benefit becoming more pronounced as compute increases. At the smallest budgets, however, the second-lowest activation ratio slightly outperforms the sparsest setting. This exception is consistent with the intuition that extremely sparse routing exposes a larger routed expert pool, which may require sufficient training compute to be effectively optimized. The trend suggests that MoE dLLMs can increasingly exploit this expert capacity at larger scales.

\findingheadline{Expert granularity: $G=8$--$16$ is robust.}
We next examine expert granularity \(G\) in Figure~\ref{fig:architecture}(\subref{fig:architecture-granularity}). \(G\) does not show a monotonic trend across compute scales, suggesting that it is not a primary scaling direction. Instead, it mainly reflects a trade-off between routing diversity and per-expert capacity: coarser experts provide stronger individual transformations but fewer routing choices, while finer-grained experts increase routing choices at the cost of narrower experts. For MoE dLLMs, this trade-off is relevant because masked denoising requires both diverse specialization over corrupted contexts and sufficient expert expressiveness. Empirically, \(G=8\) to \(G=16\) provides a robust range in our sweep, although the exact optimum does not vary systematically with compute.

\findingheadline{Shared-expert ratio: $S=33.3\%$ remains optimal.}
We examine the shared-expert ratio \(S\) in Figure~\ref{fig:architecture}(\subref{fig:architecture-shared}). Across all compute scales, the loss curves are U-shaped and reach their minima at \(S=33.3\%\). This optimum contrasts with AR MoE designs: DeepSeekMoE adopts \(S=25\%\), Qwen3 uses no shared experts~\citep{dai2024deepseekmoe,yang2025qwen3}, and \citet{tian2026towards} report a decreasing optimal ratio from \(16.7\%\) to \(8.3\%\), motivating a fixed ``one shared expert'' rule. Our dLLM sweeps instead favor a shared pathway whose capacity grows proportionally with the activated model budget, rather than a fixed shared component whose relative contribution diminishes with scale. We therefore derive a dLLM-specific rule distinct from the AR heuristic: maintain approximately one unit of shared activated capacity for every two units of routed activated capacity.

Our architecture sweeps translate the aggregate budget \(M^*(C)\) into dLLM-specific architectural choices for routing over corrupted states that vary with noise level and mask pattern rather than causal prefixes. At fixed \(M^*(C)\), larger scales favor sparser activation, while moderate expert granularity and a stable shared-capacity fraction remain robust across the studied scales.

\section{Training Large-Scale MoE dLLMs}

Guided by the scaling laws derived above, we design LLaDA MoE v2, a 30B-A3B MoE dLLM, and train it on 23.5T tokens. We briefly introduce its architecture and training strategies below, with additional details provided in Appendix~\ref{app:large-scale-details}.

\textbf{Model.}
LLaDA MoE v2 has 30B total parameters, with 3B parameters activated per token. Based on our scaling analysis, we adopt \((A,G,S)=(9.09\%,8,33.3\%)\). Each layer comprises 128 fine-grained routed experts with top-8 routing and one shared expert of width \(4d_{\mathrm{expert}}\), corresponding to \(n_s=4\) and yielding \(A=(8+4)/(128+4)=9.09\%\) and \(S=4/(8+4)=33.3\%\).

\textbf{Training strategy.}
We train LLaDA MoE v2 in five stages. Stages 1 and 2 each use 10T tokens, followed by 2T tokens of annealing in Stage 3. In Stage 4, we increase the RoPE base from 10,000 to 500,000, extend the context length from 4K to 32K, and continue training for 500B tokens. Stage 5 concludes with 1T tokens of long-context annealing, yielding the final pretrained base model.

\subsection{Benchmark Results}

\begin{table*}[t]
    \centering
    \small
    \caption{\textbf{Benchmark results.} We report results for LLaDA MoE v2, our 30B-A3B dLLM trained from scratch, alongside the representative dLLMs and AR model Qwen3. CPT denotes continued pretraining from AR models. The symbol $^*$ denotes results reported in Qwen3~\citep{yang2025qwen3}, $^\dagger$ denotes results reported in LLaDA MoE 7B-A1B~\citep{zhu2025llada}.}
    \label{tab:pretrain-results}
    \begin{tabular}{l|>{\centering\arraybackslash}m{1.42cm}>{\centering\arraybackslash}m{1.42cm}>{\centering\arraybackslash}m{1.42cm}>{\centering\arraybackslash}m{1.42cm}>{\centering\arraybackslash}m{1.42cm}|>{\centering\arraybackslash}m{1.42cm}}
        \toprule
                            & \shortstack{LLaDA\\MoE v2} & SDAR Sci & \shortstack{LLaDA\\MoE} & Dream 7B  & LLaDA 8B & Qwen3 \\
        \midrule
        Architecture        & MoE              & MoE          & MoE              & Dense        & Dense     & MoE      \\
        Modeling            & Diffusion        & Diffusion    & Diffusion        & Diffusion    & Diffusion & AR       \\
        Method              & Pretrain         & CPT          & Pretrain         & CPT          & Pretrain  & Pretrain \\
        \# Total Params     & 30B              & 30B          & 7B               & 7B           & 8B        & 30B      \\
        \# Activated Params & 3B               & 3B           & 1B               & 7B           & 8B        & 3B       \\
        \# Trained Tokens   & 23.5T            & 36 + 1.05T   & 21T              & 18 + 0.58T   & 2.3T      & 36T      \\
        \midrule
        \multicolumn{7}{c}{\textit{General Tasks}} \\
        \midrule
        MMLU                & 78.01            & 82.72        & 64.59$^\dagger$ & 69.50$^\dagger$ & 65.90$^\dagger$ & 81.38$^*$ \\
        MMLU-Pro            & 57.28            & 56.96        & 39.16$^\dagger$ & 48.15$^\dagger$ & 41.80$^\dagger$ & 61.49$^*$ \\
        CEval               & 76.11            & 86.95        & 65.56$^\dagger$ & 59.18$^\dagger$ & 70.50$^\dagger$ & 87.50     \\
        CMMLU               & 77.99            & 85.82        & 65.65$^\dagger$ & 60.87$^\dagger$ & 69.90$^\dagger$ & 86.35     \\
        HellaSwag           & 77.19            & 56.38        & 65.46           & 74.37           & 70.82           & 77.92     \\
        KorBench            & 45.92            & 40.08        & 31.20$^\dagger$ & 37.44$^\dagger$ & 33.68$^\dagger$ & 44.96     \\
        \midrule
        \multicolumn{7}{c}{\textit{Reasoning Tasks}} \\
        \midrule
        GSM8K               & 83.93            & 86.13        & 66.41$^\dagger$ & 77.79$^\dagger$ & 70.70$^\dagger$ & 91.81$^*$ \\
        MATH                & 54.72            & 48.52        & 36.10$^\dagger$ & 39.60$^\dagger$ & 27.30$^\dagger$ & 59.04$^*$ \\
        OlympiadBench       & 28.74            & 24.44        & 10.07$^\dagger$ & 10.22$^\dagger$ & 6.85$^\dagger$  & 30.96     \\
        \midrule
        \multicolumn{7}{c}{\textit{Coding Tasks}} \\
        \midrule
        CRUXEval            & 50.62            & 53.00        & 38.94           & 40.31           & 36.38           & 56.88     \\
        MBPP                & 71.00            & 60.40        & 52.40$^\dagger$ & 56.20$^\dagger$ & 38.20$^\dagger$ & 74.40$^*$ \\
        MultiPL-E           & 53.78            & 33.66        & 41.13$^\dagger$ & 27.60$^\dagger$ & 23.61$^\dagger$ & 66.53$^*$ \\
        HumanEval           & 50.00            & 33.54        & 45.73$^\dagger$ & 57.90$^\dagger$ & 33.50$^\dagger$ & 52.44     \\
        LiveCodeBench v6    & 31.86            & 39.87        & 16.18$^\dagger$ & 14.87$^\dagger$ & 2.53$^\dagger$  & 49.18     \\
        BigCodeBench        & 41.84            & 33.86        & 21.23$^\dagger$ & 18.33$^\dagger$ & 13.42$^\dagger$ & 45.70     \\
        \bottomrule
    \end{tabular}
\end{table*}

As shown in Table~\ref{tab:pretrain-results}, we compare LLaDA MoE v2 30B-A3B with five representative dLLM and AR baselines. The closest scale-matched baselines are SDAR Sci and Qwen3 30B-A3B: the former is obtained by continued pretraining from Qwen3~\citep{cheng2026sdar}, whereas the latter is a strong AR MoE model~\citep{yang2025qwen3}. LLaDA MoE v2 is trained from scratch on 23.5T tokens, 63\% of SDAR Sci's 37.05T and 65\% of Qwen3's 36T.

Across all 15 benchmarks, LLaDA MoE v2 achieves the highest average among the evaluated dLLMs (58.60), outperforming SDAR Sci by 3.78 points and the smaller dLLM baselines by at least 12.44 points. Its advantage over SDAR Sci is particularly pronounced on coding benchmarks such as HumanEval (\(+16.46\)) and BigCodeBench (\(+7.98\)), despite being pretrained from scratch rather than initialized from an AR checkpoint. Compared with Qwen3, LLaDA MoE v2 shows larger gaps on the Chinese knowledge benchmarks CEval and CMMLU and on some coding tasks, but remains close on several reasoning and coding benchmarks, including OlympiadBench (\(-2.22\)) and HumanEval (\(-2.44\)). This performance with 65\% as many pretraining tokens as Qwen3 supports the effectiveness of our scaling-law-guided design.

\begin{figure}[t]
    \centering
    \includegraphics[width=\linewidth, clip]{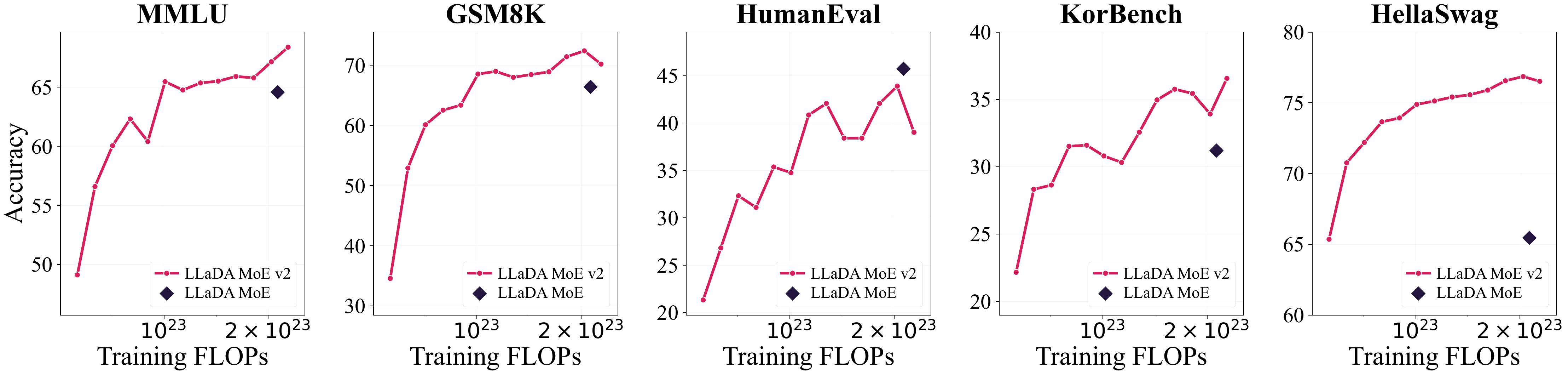}
    \vspace{-0.6em}
    \caption{\textbf{Benchmark performance versus training compute.}
    Red curves show the scaling-law-guided LLaDA MoE v2 30B-A3B model evaluated with varying training-token budgets, and the black diamond denotes the LLaDA MoE 7B-A1B model trained without scaling-law guidance.}
    \label{fig:benchmark_scaling_comparison}
\end{figure}

To further evaluate the practical value of our scaling-law-guided design, Figure~\ref{fig:benchmark_scaling_comparison} compares the benchmark performance of LLaDA MoE v2 30B-A3B across different training FLOPs with that of LLaDA MoE 7B-A1B, an MoE dLLM developed without scaling-law guidance. Across diverse benchmarks spanning knowledge, math, and code, the 30B-A3B model achieves performance comparable to or better than the 7B-A1B model at substantially lower compute budgets. Specifically, our model surpasses LLaDA MoE 7B-A1B on MMLU, GSM8K, and KorBench at approximately 50\% of its training FLOPs, and even surpasses it on HellaSwag with less than 10\% of the training FLOPs. The benchmark results and compute-controlled comparison provide practical evidence that our scaling laws offer actionable guidance for large-scale MoE dLLM design, improving the efficiency with which pretraining compute is translated into downstream performance.

\subsection{Supervised Fine-Tuning}

\begin{table}[t]
    \centering
    \small
    \setlength{\tabcolsep}{3pt}
    \renewcommand{\arraystretch}{1.08}
    \definecolor{SFTBest}{HTML}{C6D9F1}
    \definecolor{SFTSecond}{HTML}{E8F0F8}
    \caption{\textbf{SFT results.} Comparison of the LLaDA MoE v2 30B-A3B instruct model with Qwen3 30B-A3B and SDAR Chat 30B-A3B on reasoning and code benchmarks. Values marked with $^*$ are reported in the original Qwen3 paper, and marked with $^\dagger$ are reported in the original SDAR paper.}
    \label{tab:sft-results}
    \begin{tabular*}{\linewidth}{@{\extracolsep{\fill}}lcccccccc@{}}
        \toprule
        & \multicolumn{4}{c}{Reasoning} & \multicolumn{4}{c}{Code} \\
        \cmidrule(lr){2-5}\cmidrule(lr){6-9}
        Model & Math & Olympiad & AIME24 & AIME25 & MBPP & LCB v6 & BigCode & MultiPL-E \\
        \midrule
        Qwen3               & 89.80$^*$         & 57.26 & 32.80$^*$         & 21.60$^*$         & 85.48             & 31.50             & 41.14 & 66.60 \\
        SDAR Chat           & 77.80$^\dagger$   & 34.93 & 16.70$^\dagger$   & 10.80$^\dagger$   & 71.60$^\dagger$   & 21.70$^\dagger$   & 39.39 & 45.00 \\
        \textbf{LLaDA MoE v2} & 80.02           & 46.44 & 30.00             & 20.00             & 81.03             & 27.75             & 35.53 & 67.52 \\
        \bottomrule
    \end{tabular*}
\end{table}

The preceding experiments evaluate the capabilities of the pretrained LLaDA MoE v2 base model. We further examine whether this scaling-guided pretrained model can be effectively adapted to instruction following and complex reasoning through standard supervised fine-tuning (SFT). Starting from the LLaDA MoE v2 30B-A3B base checkpoint, we construct an instruct model and evaluate it on mathematical reasoning and code generation benchmarks.

We fine-tune LLaDA MoE v2 for three epochs on 7M instruction--response examples. For each example, we keep the prompt uncorrupted, apply the masking process in Equation~\ref{eq:dllm-objective} only to the response, and compute the denoising loss over the masked positions. We use a batch size of 512 sequences and a peak learning rate of \(5.0\times10^{-6}\), with linear warmup over the first 8\% of steps. We do not apply reinforcement learning (RL) after SFT and leave its integration to future work.

We compare the resulting model with Qwen3 30B-A3B no think and SDAR Chat 30B-A3B. As shown in Table~\ref{tab:sft-results}, LLaDA MoE v2 outperforms SDAR Chat on all four reasoning benchmarks and three of the four code benchmarks. Despite using fewer pretraining tokens, without the additional RL used by Qwen3~\citep{yang2025qwen3}, LLaDA MoE v2 remains close to Qwen3 on AIME 24/25~\citep{aime}, MBPP, and LiveCodeBench, and surpasses it on MultiPL-E. These results show that LLaDA MoE v2 acquires strong reasoning and coding capabilities through SFT alone.

\section{Related Work}

\textbf{Diffusion Language Models (dLLMs)} have recently emerged as a new paradigm for language modeling~\citep{austin2021structured,campbell2022continuous,chen2022analog,gulrajani2023likelihood,he2023diffusionbert,lou2023discrete,shi2024simplified,xue2024unifying,zheng2025masked,li2025survey,nie2026large,song2025seed,labs2025mercury}. In particular, masked discrete diffusion models are viewed as potential alternatives to autoregressive (AR) models, as they can generate multiple tokens in parallel at each denoising step~\citep{arriola2025block,wei2025accelerating,chen2026dflash,cheng2026dspark}. Recent work has scaled dLLMs both by pretraining from scratch~\citep{nie2026large,zhu2025llada, zhu2026llada} and by adapting pretrained AR models~\citep{gong2025scaling,dream2025,bie2025llada2}. Meanwhile, several studies have begun to characterize the scaling behavior of dLLMs~\citep{nie2025scaling,ni2025training,von2026scaling,ni2025diffusion,sahoo2026scaling}, though most of them are confined to dense architectures.

\textbf{Mixture-of-Experts (MoE) and Scaling Laws.} By replacing a single feed-forward network with multiple fine-grained experts and selectively activating only a subset of parameters for each token, MoE architectures enable models to increase capacity without a proportional increase in computation~\citep{shazeer2017outrageously,lepikhin2020gshard,fedus2022switch,du2022glam,jiang2024mixtral,dai2024deepseekmoe}. Building on scaling law research for language models~\citep{kaplan2020scaling,hoffmann2022training}, recent studies have used scaling insights to guide the design of compute-efficient MoE architectures~\citep{clark2022unified,abnar2025parameters,ludziejewski2025joint}, and several works have also applied MoE to dLLMs~\citep{zhu2025llada,feng2026dmoe}. Existing MoE dLLMs, however, largely adopt architectural choices developed for AR models, leaving their optimization, compute allocation, and expert architecture scaling behavior largely uncharacterized.

\section{Conclusion}

In this work, we characterize how optimization hyperparameters, model--data allocation, and expert architecture scale for MoE dLLMs, finding that AR trends provide useful priors but require dLLM-specific calibration. These results yield practical design principles that guide the from-scratch training of LLaDA MoE v2 30B-A3B. The model approaches Qwen3 on several benchmarks with fewer pretraining tokens, while SFT alone, without RL, produces an instruct model that outperforms SDAR Chat on seven of eight reasoning and coding tasks. Our experiments vary the scaling dimensions separately and therefore do not capture their interactions. Nevertheless, the large-scale results demonstrate the practical value of scaling-law-guided MoE dLLM design.

\bibliography{iclr2026_conference}
\bibliographystyle{iclr2026_conference}

\newpage

\appendix

\section{Scaling Laws for MoE dLLMs}
\label{app:scaling-setup}

\subsection{Compute Accounting}
\label{app:compute-accounting}

We distinguish total parameter count, activated parameter count, and the activated computation used in our scaling analysis. Let \(n_{\mathrm{layer}}\) be the number of Transformer layers, \(d_{\mathrm{model}}\) the hidden size, \(s\) the sequence length, and \(r_{\mathrm{kv}}=n_{\mathrm{kvheads}}/n_{\mathrm{heads}}\) the ratio of key-value heads to query heads. Each MoE layer contains \(n_e\) routed experts, each with intermediate width \(d_{\mathrm{expert}}\), of which \(n_a\) are selected per token. In the implemented architecture, every token is additionally processed by a single shared expert with intermediate width \(d_{\mathrm{share}}\). To express routed and shared capacity in common units, we define \(n_s \equiv d_{\mathrm{share}}/d_{\mathrm{expert}}\), or equivalently \(d_{\mathrm{share}}=n_s d_{\mathrm{expert}}\). Thus, \(n_s\) denotes the shared capacity measured in units of one routed expert, rather than the physical number of shared experts. With SwiGLU~\citep{shazeer2020glu}, the three projection matrices contain \(3d_{\mathrm{model}}d_{\mathrm{expert}}\) parameters for each routed expert and \(3d_{\mathrm{model}}d_{\mathrm{share}}=3n_s d_{\mathrm{model}}d_{\mathrm{expert}}\) parameters for the single shared expert.

Ignoring biases and normalization parameters, the total and activated non-embedding parameter counts are
\begin{align}
    P_{\mathrm{nonemb}} &= n_{\mathrm{layer}} \left[ 2d_{\mathrm{model}}^2(1+r_{\mathrm{kv}}) + d_{\mathrm{model}}n_e + 3d_{\mathrm{model}}(n_e d_{\mathrm{expert}}+d_{\mathrm{share}}) \right], \\
    P_{\mathrm{act,nonemb}} &= n_{\mathrm{layer}} \left[ 2d_{\mathrm{model}}^2(1+r_{\mathrm{kv}}) + d_{\mathrm{model}}n_e + 3d_{\mathrm{model}}(n_a d_{\mathrm{expert}}+d_{\mathrm{share}}) \right].
\end{align}
The first term accounts for the query, key, value, and output projections; the second accounts for the router, which scores all \(n_e\) routed experts for every token; and the final term accounts for the routed experts and the single shared expert. Substituting \(d_{\mathrm{share}}=n_s d_{\mathrm{expert}}\) recovers the equivalent-unit forms \(3d_{\mathrm{model}}d_{\mathrm{expert}}(n_e+n_s)\) and \(3d_{\mathrm{model}}d_{\mathrm{expert}}(n_a+n_s)\), respectively. We use an input embedding matrix and a separate LM head matrix, each containing \(Vd_{\mathrm{model}}\) parameters for vocabulary size \(V\). Their combined \(2Vd_{\mathrm{model}}\) parameters are included when reporting total or activated parameter counts but excluded from the non-embedding quantities used below.

We count one multiply--accumulate as two FLOPs. The forward FLOPs per token of one layer are approximated by
\begin{align}
    F_{\mathrm{attn}} &= 4d_{\mathrm{model}}^2(1+r_{\mathrm{kv}}) + 4s d_{\mathrm{model}}, \\
    F_{\mathrm{MoE}} &= 2d_{\mathrm{model}}n_e + 6d_{\mathrm{model}}(n_a d_{\mathrm{expert}}+d_{\mathrm{share}}),
\end{align}
where the two terms in \(F_{\mathrm{attn}}\) correspond to the attention projections and the two sequence-level attention matrix multiplications, respectively. The terms in \(F_{\mathrm{MoE}}\) account for routing, the selected routed experts, and the single shared expert. Approximating the backward pass as twice the forward pass, we define the activated non-embedding FLOPs per token as
\begin{equation}
    M = 3n_{\mathrm{layer}} \left[ 4d_{\mathrm{model}}^2(1+r_{\mathrm{kv}}) + 4s d_{\mathrm{model}} + 2d_{\mathrm{model}}n_e + 6d_{\mathrm{model}}(n_a d_{\mathrm{expert}}+d_{\mathrm{share}}) \right].
    \label{eq:activated-training-flops}
\end{equation}
This FLOPs accounting omits the input embedding, LM head, normalization operations, nonlinearities, and attention softmax. All Transformer layers in our models share the same architecture; accordingly, Equation~\ref{eq:activated-training-flops} multiplies the per-layer FLOPs by \(n_{\mathrm{layer}}\).

Finally, let \(D\) denote the total number of nominal tokens processed during training, including both masked and visible positions. The compute budget used throughout our scaling law experiments is
\begin{equation}
    C = MD.
\end{equation}
Masked and visible tokens incur the same Transformer computation, so the sampled corruption level changes the number of supervised prediction targets but not the accounted FLOPs.

\subsection{MoE Implementation}
\label{app:moe-implementation}

Throughout the scaling experiments, we use a MoE Transformer architecture in which every layer pairs grouped-query attention (GQA)~\citep{ainslie2023gqa} with an MoE feed-forward block comprising the routed experts and the single shared expert parameterized in Appendix~\ref{app:compute-accounting}, all implemented as SwiGLU networks; the only exceptions arise in the architecture experiments, where some configurations omit the shared expert.

Routing is performed independently for each token at every layer. Given the token representation \(h\), a linear router produces logits \(r(h)\in\mathbb{R}^{n_e}\) and routing scores \(p(h)=\operatorname{softmax}(r(h))\). Let \(\mathcal{T}(h)\) contain the indices of the \(n_a\) largest routing scores. We renormalize the selected scores as
\begin{equation}
    w_i(h) = \frac{p_i(h)}{\sum_{j\in\mathcal{T}(h)}p_j(h)}, \qquad i\in\mathcal{T}(h).
\end{equation}
The routed expert output is
\begin{equation}
    E_{\mathrm{route}}(h)
    =
    \sum_{i\in\mathcal{T}(h)} w_i(h)E_i(h).
\end{equation}
The routed and shared pathways are combined as \(E_{\mathrm{share}}(h)+\lambda E_{\mathrm{route}}(h)\), where \(E_i\) is a routed expert, \(E_{\mathrm{share}}\) is the shared expert, and \(\lambda\) balances the output scales of the two pathways. The routed-only configuration in the shared-expert-ratio experiments omits the shared term and does not use a scaling factor.

For configurations with \(n_s>0\), we estimate \(\lambda\) by matching the expected output norms of the shared and routed pathways at initialization, following the gate scaling heuristic of \citet{liu2025muon}. Treating the shared expert of width \(n_s d_{\mathrm{expert}}\) as \(n_s\) expert-width units, and assuming that all expert outputs have equal norms and are pairwise orthogonal at initialization, the shared pathway has norm proportional to \(\sqrt{n_s}\), whereas the unscaled routed pathway has norm proportional to \((\sum_{i\in\mathcal{T}(h)}w_i(h)^2)^{1/2}\). Approximating the initialization distribution of the router logits by \(r(h)\sim\mathcal{N}(0,I_{n_e})\), we estimate
\begin{equation}
    \lambda = \mathbb{E}_{r(h)\sim\mathcal{N}(0,I_{n_e})}
    \left[
        \frac{\sqrt{n_s}}{\left(\sum_{i\in\mathcal{T}(h)}w_i(h)^2\right)^{1/2}}
    \right],
    \label{eq:moe-routed-scaling}
\end{equation}
where the expectation is approximated by Monte Carlo sampling independently for each architecture configuration with a shared expert.

We train the router jointly with the rest of the model. The training objective is
\begin{equation}
    \mathcal{L} = \mathcal{L}_{\mathrm{diff}} + \alpha_{\mathrm{aux}}\mathcal{L}_{\mathrm{bal}} + \alpha_{z}\mathcal{L}_{z},
\end{equation}
where \(\mathcal{L}_{\mathrm{diff}}\) is the denoising objective in Equation~\ref{eq:dllm-objective}, \(\mathcal{L}_{\mathrm{bal}}\) is the load balancing auxiliary loss, and \(\mathcal{L}_{z}\) is the router \(z\)-loss; following commonly used MoE training settings, we set \(\alpha_{\mathrm{aux}}=0.01\) and \(\alpha_z=0.001\)~\citep{fedus2022switch,shazeer2017outrageously,zoph2022st}. Both auxiliary terms are added directly to the denoising loss, and their coefficients are held fixed across all scaling and architecture sweeps.

\subsection{Hyperparameter Scaling}
\label{app:hyperparameter-scaling}

We conduct the hyperparameter scaling experiments at three model scales---158M, 1B, and 3.6B---under compute budgets ranging from \(10^{18}\) to \(3\times10^{20}\) FLOPs. All runs use the same pretraining data and a sequence length of 4096, and optimize the denoising objective in Equation~\ref{eq:dllm-objective}. We use AdamW with \((\beta_1,\beta_2)=(0.9,0.95)\) and a weight decay of 0.1. The learning rate is linearly warmed up for 2,000 optimizer steps to the peak value \(\eta\), allowing optimization to enter a stable regime before the peak rate is held constant until the final 10\% of the training compute and then decayed to \(0.1\eta\) using a cosine schedule.

For each model scale, we jointly search the global nominal token batch size \(B\) and peak learning rate \(\eta\), while holding the architecture and training-token budget fixed within each compute setting. Table~\ref{tab:hyperparameter-configurations} summarizes the model architectures, compute budgets, and corresponding search grids.

\begin{table*}[t]
    \centering
    \small
    \setlength{\tabcolsep}{4.5pt}
    \renewcommand{\arraystretch}{1.12}
    \caption{\textbf{Configurations for the hyperparameter scaling sweep.}
    The upper block lists the model architectures; the lower block reports the compute budgets \(C\) (in FLOPs) and joint search grids for the global nominal token batch size \(B\) and peak learning rate \(\eta\).}
    \label{tab:hyperparameter-configurations}
    \begin{tabular*}{\textwidth}{@{\extracolsep{\fill}}lcccccccc@{}}
        \toprule
        \multicolumn{9}{c}{Model architecture} \\
        \cmidrule(lr){1-9}
        Model scale
        & \(n_{\mathrm{layer}}\)
        & \(d_{\mathrm{model}}\)
        & \(n_{\mathrm{heads}}\)
        & \(n_{\mathrm{kvheads}}\)
        & \(n_e\)
        & \(n_a\)
        & \(n_s\)
        & \(d_{\mathrm{expert}}\) \\
        \midrule
        158M & 6  & 256  & 8  & 2 & 64 & 4 & 1 & 256  \\
        1B   & 10 & 640  & 10 & 2 & 64 & 4 & 1 & 640  \\
        3.6B & 16 & 1024 & 16 & 4 & 64 & 4 & 1 & 1024 \\
        \addlinespace[0.45em]
        \multicolumn{9}{c}{Search configuration} \\
        \cmidrule(lr){1-9}
        Model scale
        & \multicolumn{3}{c}{\(C\)}
        & \multicolumn{2}{c}{\(B\)}
        & \multicolumn{3}{c}{\(\eta\)} \\
        \midrule
        158M
        & \multicolumn{3}{c}{\(\{1,2,3,6,8\}\times10^{18}\)}
        & \multicolumn{2}{c}{\(\{2^{17},2^{18},2^{19},2^{20}\}\)}
        & \multicolumn{3}{c}{\(\{1,1.4,2,2.8\}\times10^{-3}\)} \\
        1B
        & \multicolumn{3}{c}{\(\{1,2,3,6,8,10\}\times10^{19}\)}
        & \multicolumn{2}{c}{\(\{2^{18},2^{19},2^{20},2^{21}\}\)}
        & \multicolumn{3}{c}{\(\{0.7,1,1.4,2\}\times10^{-3}\)} \\
        3.6B
        & \multicolumn{3}{c}{\(\{0.8,1,2,3\}\times10^{20}\)}
        & \multicolumn{2}{c}{\(\{2^{21},2^{22},2^{23}\}\)}
        & \multicolumn{3}{c}{\(\{1,4,7\}\times10^{-4}\)} \\
        \bottomrule
    \end{tabular*}
\end{table*}

For each run, we define its loss as the average training loss over the final \(0.5\%\) of its allocated training FLOPs. At each compute budget, we identify the minimum average loss and regard configurations whose losses are no more than \(0.25\%\) above this minimum as near-optimal~\citep{bi2024deepseek}. We include all near-optimal configurations in log--log linear regressions of batch size and learning rate against \(C\), yielding the final scaling laws.

\subsection{Compute Allocation Scaling}
\label{app:compute-allocation}

We conduct IsoFLOP sweeps under compute budgets ranging from \(10^{17}\) to \(10^{20}\) FLOPs. All runs use the same pretraining data, denoising objective, and optimizer as the hyperparameter scaling experiments in Appendix~\ref{app:hyperparameter-scaling}. At each compute budget \(C\), we set the nominal token batch size and peak learning rate according to the fitted hyperparameter scaling laws reported in the main text. Tables~\ref{tab:compute-allocation-architectures} and~\ref{tab:compute-allocation-configurations} report the candidate model architectures and their corresponding allocation configurations, respectively. Because the number of optimizer steps varies across model--data allocations, we adapt the warmup length to each run. Let \(T\) denote the total number of optimizer steps implied by the allocation. We linearly warm up the learning rate to its peak over \(T_{\mathrm{warm}}=\max(0.01T,100)\) steps, maintain the peak rate until the final \(10\%\) of the training steps, and then decay it to \(10\%\) of the peak using a cosine schedule. The 100-step floor provides a minimum warmup period for optimization stability~\citep{nie2025scaling}.

For each compute budget, we evaluate a set of MoE models spanning different amounts of activated model-side computation. We measure the model side by the activated non-embedding FLOPs per token \(M\), and assign each model a training-token budget \(D=C/M\). At each compute budget, we select the evaluated allocation with the lowest loss as the empirical optimal allocation point. We fit the selected model-side optima using linear regression in log--log space and derive the corresponding data-side frontier from \(D^*(C)=C/M^*(C)\), yielding the final compute-allocation scaling laws reported in the main text.

\subsection{MoE Architecture Scaling}
\label{app:architecture-scaling}

We conduct the MoE architecture sweeps at five reference compute budgets, \(C\in\{6\times10^{17},2\times10^{18},6\times10^{18},2\times10^{19},6\times10^{19}\}\) FLOPs. At each reference budget, we keep the Transformer backbone fixed and vary one architectural dimension at a time while holding the other two fixed as closely as permitted by the discrete configurations. The sweep grids are parameterized using the definitions in Section~\ref{sec:preliminaries}: the activation ratio \(A=(n_a+n_s)/(n_e+n_s)\), the expert granularity \(G=2d_{\mathrm{model}}/d_{\mathrm{expert}}\), and the shared-expert ratio \(S=n_s/(n_a+n_s)\). We evaluate the fitted compute-allocation scaling laws at the reference budget \(C\) to obtain the target activated non-embedding FLOPs per token \(M^*(C)\) and the compute-optimal token count \(D^*(C)\). To make architecture selection representative of the overtraining regime commonly used in large-scale pretraining~\citep{grattafiori2024llama, gadre2025language, tian2026towards}, we train each candidate for \(3D^*(C)\) nominal tokens, corresponding to approximately \(3C\) training FLOPs. We therefore evaluate the fitted hyperparameter scaling laws at the actual training budget \(3C\) and set the nominal token batch size \(B\) and peak learning rate \(\eta\) accordingly. The common allocation targets and training hyperparameters are reported in Table~\ref{tab:architecture-training-configurations}; the exact model architectures used in the three sweeps are reported in Tables~\ref{tab:architecture-activation-configurations}, \ref{tab:architecture-granularity-configurations}, and~\ref{tab:architecture-shared-configurations}. All remaining training settings are identical to those of the compute-allocation sweeps in Appendix~\ref{app:compute-allocation}.

Within each sweep, the candidate architectures are constructed so that, ignoring the negligible router contribution, the activated model-side budget \(M^*(C)\) is preserved. In the activation-ratio sweep, we fix the Transformer backbone, \(n_a\), \(n_s\), and \(d_{\mathrm{expert}}\), and vary only the number of routed experts \(n_e\): the activated expert computation per token is unchanged, while the total parameter count grows as \(A\) decreases. In the expert-granularity sweep, we vary \(d_{\mathrm{expert}}\) and scale \(n_e\), \(n_a\), and \(n_s\) in inverse proportion, preserving \(A\), \(S\), and both the routed and shared activated widths, \(n_a d_{\mathrm{expert}}\) and \(n_s d_{\mathrm{expert}}\). In the shared-expert-ratio sweep, we fix \(n_e\) and \(d_{\mathrm{expert}}\) and redistribute the fixed activated expert width \((n_a+n_s)d_{\mathrm{expert}}\) between the shared and routed pathways by varying \(n_s\) and \(n_a\); since \(n_s\) enters the denominator of \(A\), this redistribution induces a slight drift in the activation ratio across candidates, which is negligible and does not affect the controlled comparison.

For each run, we define its loss as the average training loss over the final \(0.5\%\) of its allocated training FLOPs, as in Appendix~\ref{app:hyperparameter-scaling}. At each reference budget, we select the candidate with the lowest loss along each architectural dimension.

\section{Training Large-Scale MoE dLLMs}
\label{app:large-scale-details}

\subsection{Model Architecture}
\label{app:large-scale-architecture}

We train LLaDA MoE v2 30B-A3B, a large MoE dLLM whose detailed architecture is reported in Table~\ref{tab:large-scale-architecture}. The routing rule, the combination of routed and shared expert outputs, and the auxiliary objectives follow Appendix~\ref{app:moe-implementation}.

The Transformer backbone contains 32 layers with hidden size 3072 and uses grouped-query attention (GQA) with 32 query heads and 4 key-value heads. The model uses a vocabulary of 157,184 tokens. All Transformer layers use an MoE feed-forward block.

To realize the architecture-scaling recommendations, we use \(n_e=128\) routed experts, activate \(n_a=8\) of them per token, and allocate \(n_s=4\) expert-width units to the single shared expert. Setting \(G=8\) gives \(d_{\mathrm{expert}}=d_{\mathrm{model}}/4\) and hence \(d_{\mathrm{share}}=4d_{\mathrm{expert}}=d_{\mathrm{model}}\). This discrete configuration yields \(A=(8+4)/(128+4)=9.09\%\) and \(S=4/(8+4)=33.3\%\).

\begin{table*}[t]
    \centering
    \small
    \setlength{\tabcolsep}{4pt}
    \renewcommand{\arraystretch}{1.1}
    \caption{\textbf{Architecture of LLaDA MoE v2 30B-A3B.}}
    \label{tab:large-scale-architecture}
    \begin{tabular*}{\textwidth}{@{\extracolsep{\fill}}ccccccccc@{}}
        \toprule
        \multicolumn{2}{c}{Parameter scale}
        & \multicolumn{4}{c}{Transformer backbone}
        & \multicolumn{3}{c}{MoE configuration} \\
        \cmidrule(lr){1-2}\cmidrule(lr){3-6}\cmidrule(lr){7-9}
        Total & Activated & \(n_{\mathrm{layer}}\) & \(d_{\mathrm{model}}\) & \(n_{\mathrm{heads}}\) & \(n_{\mathrm{kvheads}}\) & \(n_e\) & \(n_a\) & \(n_s\) \\
        \midrule
        30.6B   & 3.4B  & 32    & 3072  & 32    & 4 & 128   & 8 & 4 \\
        \bottomrule
    \end{tabular*}
\end{table*}

\subsection{Pretraining}
\label{app:pretraining-details}

We pretrain LLaDA MoE v2 30B-A3B from scratch for a total of 23.5T nominal tokens. The pretraining corpus is constructed from a broad collection of high quality text gathered from the web. We apply a standard data processing pipeline that collects the raw text, removes boilerplate and malformed or low quality documents, deduplicates repeated content, and filters harmful material. Throughout pretraining, we optimize the denoising objective in Equation~\ref{eq:dllm-objective}; the routing procedure and auxiliary objectives follow Appendix~\ref{app:moe-implementation}.

The five-stage data schedule is summarized in Table~\ref{tab:large-scale-pretraining-stages}. Stages 1 and 2 draw separate 10T-token samples from the same source corpus, with the Stage 2 mixture assigning slightly more weight to mathematical reasoning and code data. For Stage 3, we construct a curated 1T-token annealing corpus after cleaning, deduplication, and harmful-content filtering, and train on it for two epochs, yielding 2T training tokens. Stages 4 and 5 primarily use long-form data with sequence lengths of up to 32K tokens.

We train in BF16 precision using AdamW with \((\beta_1,\beta_2)=(0.9,0.95)\) and a weight decay of 0.1~\citep{loshchilov2017decoupled}. The global nominal token batch size is 33,554,432 tokens. Each stage uses a separate learning-rate schedule that linearly warms up from zero to its stage-specific peak over the first 2,000 optimizer steps. During the final 10\% of the allocated training compute in Stages 1--4, the learning rate is decayed to the peak rate of the following stage using a cosine schedule; in Stage 5, it is decayed to \(5.0\times10^{-6}\) using the same schedule. The peak learning rates for Stages 1--5 are \(1.5\times10^{-4}\), \(1.0\times10^{-4}\), \(5.0\times10^{-5}\), \(1.0\times10^{-5}\), and \(7.0\times10^{-6}\), respectively. At the transition from Stage 3 to Stage 4, we increase the RoPE base from 10,000 to 500,000 to extend the context length from 4K to 32K~\citep{su2024roformer,xiong2024effective}. The full pretraining run consumed approximately 460,000 NVIDIA B200 GPU hours.

\begin{table*}[t]
    \centering
    \small
    \setlength{\tabcolsep}{7pt}
    \renewcommand{\arraystretch}{1.1}
    \caption{\textbf{Five-stage pretraining schedule for LLaDA MoE v2 30B-A3B.}}
    \label{tab:large-scale-pretraining-stages}
    \begin{tabular*}{\textwidth}{@{\extracolsep{\fill}}clccc@{}}
        \toprule
        Stage & Training phase & Tokens & Context length & RoPE base \\
        \midrule
        1 & Base pretraining 1     & 10T  & 4K  & 10,000  \\
        2 & Base pretraining 2     & 10T  & 4K  & 10,000  \\
        3 & Pretraining annealing  & 2T   & 4K  & 10,000  \\
        4 & Context extension      & 500B & 32K & 500,000 \\
        5 & Long-context annealing & 1T   & 32K & 500,000 \\
        \bottomrule
    \end{tabular*}
\end{table*}

\subsection{Supervised Fine-Tuning}
\label{app:sft-details}

Starting from the final pretrained checkpoint, we fine-tune LLaDA MoE v2 30B-A3B for three epochs on 7M instruction--response examples, primarily comprising single-turn mathematical reasoning and code generation tasks. We process the data following the same general procedure used for pretraining, removing malformed or low quality examples, deduplicating repeated content, and filtering harmful material, and then format every example using a unified conversation template. The formatted examples are packed into non-overlapping 8K-token training sequences.

For each instruction--response pair, we concatenate the prompt and response but apply the forward masking process in Equation~\ref{eq:dllm-objective} only to response tokens, leaving the prompt available as uncorrupted conditioning context and computing the denoising loss only at masked response positions. The MoE load balancing loss and router \(z\)-loss described in Appendix~\ref{app:moe-implementation} remain active during SFT, with their coefficients unchanged at \(\alpha_{\mathrm{aux}}=0.01\) and \(\alpha_z=0.001\).

We update all model parameters using AdamW with \((\beta_1,\beta_2)=(0.9,0.999)\), a weight decay of 0.1, and gradient clipping at a maximum norm of 1.0. The global batch size is 512 sequences. The learning rate is linearly warmed up to \(5.0\times10^{-6}\) over the first 8\% of training steps and then follows a cosine schedule that decays it to a minimum of \(1.0\times10^{-6}\). We use the final checkpoint as the instruct model and apply no reinforcement learning stage after SFT.

\subsection{Evaluation}
\label{app:evaluation}

We evaluate LLaDA MoE v2 across a diverse suite of benchmarks covering general tasks (MMLU~\citep{hendrycks2020measuring}, MMLU-Pro~\citep{wang2024mmlu}, CEval~\citep{huang2023c}, CMMLU~\citep{li2024cmmlu}, HellaSwag~\citep{zellers2019hellaswag}, KorBench~\citep{ma2025kor}), mathematical reasoning (GSM8K~\citep{cobbe2021training}, MATH~\citep{hendrycks2021measuring}, OlympiadBench~\citep{he2024olympiadbench}), and code generation (CRUXEval~\citep{gu2024cruxeval}, MBPP~\citep{austin2021program}, MultiPL-E~\citep{cassano2022multipl}, HumanEval~\citep{chen2021evaluating}, LiveCodeBench~\citep{jain2025livecodebench}, BigCodeBench~\citep{zhuo2025bigcodebench}).

For the base-model results in Table~\ref{tab:pretrain-results}, we use conditional likelihood for multiple-choice tasks and conditional generation for the remaining tasks, whereas all results in the SFT evaluation in Table~\ref{tab:sft-results} are obtained through conditional generation. For each model, we preferentially report results from its official publication~\citep{yang2025qwen3,cheng2026sdar,zhu2025llada,dream2025,nie2026large}; when a benchmark result is unavailable, we report the score obtained under our unified evaluation configuration.

We use conditional likelihood on MMLU, MMLU-Pro, CEval, CMMLU, and HellaSwag. For each example, we compute the conditional likelihood of every candidate answer given the prompt, select the candidate with the highest likelihood, and report accuracy over the benchmark. For an AR model, we compute the conditional log-likelihood under its left-to-right factorization. For LLaDA MoE v2, we follow the likelihood-evaluation protocols used in SMDM, LLaDA, and the previous LLaDA MoE~\citep{nie2025scaling,nie2026large,zhu2025llada}. For SDAR, we estimate conditional likelihood using the method introduced in Block Diffusion~\citep{arriola2025block}.

For conditional-generation tasks, each model generates a completion from the benchmark prompt using its native generation procedure. For code tasks, we extract the code from the response and execute it against the benchmark test cases. For mathematical reasoning tasks, we extract the final answer and determine correctness using the equivalence checker.

All instruct models are evaluated through conditional generation. We allow at most 1,024 generated tokens on each benchmark. Because this limit can be insufficient for models to produce a final answer on MATH, OlympiadBench, AIME 2024, and AIME 2025, we increase the limit to 4,096 tokens for these benchmarks. LLaDA MoE v2 uses semi-autoregressive sampling~\citep{nie2026large} with a block size of 64 and a total number of denoising steps equal to the generation length. For SDAR, we follow its recommended block diffusion decoding procedure with a block size of 4 and likewise set the total number of sampling steps to the generation length~\citep{cheng2026sdar}.

\begin{table}[H]
    \centering
    \small
    \setlength{\tabcolsep}{3.5pt}
    \renewcommand{\arraystretch}{1.12}
    \caption{\textbf{Model architectures for the compute-allocation sweeps.}
    Each row specifies a candidate model architecture.}
    \label{tab:compute-allocation-architectures}
    \begin{tabular*}{\textwidth}{@{\extracolsep{\fill}}lcccccccc@{}}
        \toprule
        Model scale
        & \(n_{\mathrm{layer}}\)
        & \(d_{\mathrm{model}}\)
        & \(n_{\mathrm{heads}}\)
        & \(n_{\mathrm{kvheads}}\)
        & \(n_e\)
        & \(n_a\)
        & \(n_s\)
        & \(d_{\mathrm{expert}}\) \\
        \midrule
        60M  & 6  & 128  & 8  & 2 & 64 & 4 & 1 & 128  \\
        63M  & 7  & 128  & 8  & 2 & 64 & 4 & 1 & 128  \\
        66M  & 8  & 128  & 8  & 2 & 64 & 4 & 1 & 128  \\
        69M  & 9  & 128  & 8  & 2 & 64 & 4 & 1 & 128  \\
        103M & 6  & 192  & 8  & 2 & 64 & 4 & 1 & 192  \\
        111M & 7  & 192  & 8  & 2 & 64 & 4 & 1 & 192  \\
        119M & 8  & 192  & 8  & 2 & 64 & 4 & 1 & 192  \\
        184M & 8  & 256  & 8  & 2 & 64 & 4 & 1 & 256  \\
        242M & 7  & 320  & 8  & 2 & 64 & 4 & 1 & 320  \\
        296M & 6  & 384  & 8  & 2 & 64 & 4 & 1 & 384  \\
        325M & 7  & 384  & 8  & 2 & 64 & 4 & 1 & 384  \\
        354M & 8  & 384  & 8  & 2 & 64 & 4 & 1 & 384  \\
        472M & 6  & 512  & 8  & 2 & 64 & 4 & 1 & 512  \\
        574M & 6  & 576  & 8  & 2 & 64 & 4 & 1 & 576  \\
        575M & 8  & 512  & 8  & 2 & 64 & 4 & 1 & 512  \\
        627M & 9  & 512  & 8  & 2 & 64 & 4 & 1 & 512  \\
        731M & 11 & 512  & 8  & 2 & 64 & 4 & 1 & 512  \\
        1B   & 10 & 640  & 10 & 2 & 64 & 4 & 1 & 640  \\
        1.1B & 11 & 640  & 10 & 2 & 64 & 4 & 1 & 640  \\
        1.4B & 10 & 768  & 12 & 4 & 64 & 4 & 1 & 768  \\
        1.6B & 12 & 768  & 12 & 4 & 64 & 4 & 1 & 768  \\
        1.8B & 14 & 768  & 12 & 4 & 64 & 4 & 1 & 768  \\
        2.8B & 12 & 1024 & 16 & 4 & 64 & 4 & 1 & 1024 \\
        3.2B & 14 & 1024 & 16 & 4 & 64 & 4 & 1 & 1024 \\
        7.5B & 22 & 1280 & 20 & 4 & 64 & 4 & 1 & 1280 \\
        \bottomrule
    \end{tabular*}
\end{table}

\newpage

\begingroup
\small
\setlength{\tabcolsep}{6pt}
\renewcommand{\arraystretch}{1.12}
\setlength{\LTleft}{0pt plus 1fill}
\setlength{\LTright}{0pt plus 1fill}
\setlength{\LTcapwidth}{\textwidth}
\begin{longtable}{@{}*{5}{
    >{\centering\arraybackslash}
    p{\dimexpr0.2\textwidth-1.6\tabcolsep\relax}
    }@{}}
    \caption{\textbf{Allocation configurations for the compute-allocation sweeps.}
    Each row reports the compute budget \(C\) (in FLOPs), model scale, training-token budget \(D\), global nominal token batch size \(B\), and peak learning rate \(\eta\).}
    \label{tab:compute-allocation-configurations} \\
    \toprule
    \(C\)
    & Model scale
    & \(D\;(\times 10^9)\)
    & \(B\)
    & \(\eta\) \\
    \midrule
    \endfirsthead
    \multicolumn{5}{c}{\tablename~\thetable{} continued} \\
    \toprule
    \(C\)
    & Model scale
    & \(D\;(\times 10^9)\)
    & \(B\)
    & \(\eta\) \\
    \midrule
    \endhead
    \multicolumn{5}{r}{Continued on next page} \\
    \endfoot
    \bottomrule
    \endlastfoot

    \multirow{8}{*}{\(10^{17}\)}
        & 60M  & 2.07  & \multirow{8}{*}{\(2^{18}\)} & \multirow{8}{*}{\(4.4\times10^{-3}\)} \\*
        & 63M  & 1.77  & & \\*
        & 66M  & 1.55  & & \\*
        & 69M  & 1.38  & & \\*
        & 103M & 1.25  & & \\*
        & 111M & 1.07  & & \\*
        & 119M & 0.934 & & \\*
        & 184M & 0.639 & & \\
    \midrule
    \multirow{9}{*}{\(3\times10^{17}\)}
        & 63M  & 5.32  & \multirow{9}{*}{\(2^{19}\)} & \multirow{9}{*}{\(3.4\times10^{-3}\)} \\*
        & 66M  & 4.65  & & \\*
        & 69M  & 4.14  & & \\*
        & 103M & 3.74  & & \\*
        & 111M & 3.20  & & \\*
        & 119M & 2.80  & & \\*
        & 184M & 1.92  & & \\*
        & 296M & 1.45  & & \\*
        & 472M & 0.945 & & \\
    \midrule
    \multirow{11}{*}{\(10^{18}\)}
        & 119M & 9.34 & \multirow{11}{*}{\(2^{20}\)} & \multirow{11}{*}{\(2.5\times10^{-3}\)} \\*
        & 184M & 6.39 & & \\*
        & 242M & 5.37 & & \\*
        & 296M & 4.83 & & \\*
        & 325M & 4.14 & & \\*
        & 354M & 3.62 & & \\*
        & 472M & 3.15 & & \\*
        & 574M & 2.63 & & \\*
        & 575M & 2.36 & & \\*
        & 627M & 2.10 & & \\*
        & 731M & 1.72 & & \\
    \midrule
    \multirow{11}{*}{\(3\times10^{18}\)}
        & 184M & 19.2 & \multirow{11}{*}{\(2^{20}\)} & \multirow{11}{*}{\(2.0\times10^{-3}\)} \\*
        & 242M & 16.1 & & \\*
        & 296M & 14.5 & & \\*
        & 325M & 12.4 & & \\*
        & 354M & 10.9 & & \\*
        & 472M & 9.45 & & \\*
        & 574M & 7.89 & & \\*
        & 575M & 7.09 & & \\*
        & 627M & 6.30 & & \\*
        & 731M & 5.16 & & \\*
        & 1B   & 4.03 & & \\
    \midrule
    \multirow{12}{*}{\(10^{19}\)}
        & 296M & 48.3 & \multirow{12}{*}{\(2^{21}\)} & \multirow{12}{*}{\(1.4\times10^{-3}\)} \\*
        & 325M & 41.4 & & \\*
        & 354M & 36.2 & & \\*
        & 472M & 31.5 & & \\*
        & 574M & 26.3 & & \\*
        & 575M & 23.6 & & \\*
        & 627M & 21.0 & & \\*
        & 731M & 17.2 & & \\*
        & 1B   & 13.4 & & \\*
        & 1.1B & 12.2 & & \\*
        & 1.4B & 9.94 & & \\*
        & 1.6B & 8.29 & & \\
    \midrule
    \multirow{12}{*}{\(3\times10^{19}\)}
        & 472M & 94.5 & \multirow{12}{*}{\(2^{21}\)} & \multirow{12}{*}{\(1.1\times10^{-3}\)} \\*
        & 574M & 78.9 & & \\*
        & 575M & 70.9 & & \\*
        & 627M & 63.0 & & \\*
        & 731M & 51.6 & & \\*
        & 1B   & 40.3 & & \\*
        & 1.1B & 36.6 & & \\*
        & 1.4B & 29.8 & & \\*
        & 1.6B & 24.9 & & \\*
        & 1.8B & 21.3 & & \\*
        & 2.8B & 15.5 & & \\*
        & 3.2B & 13.3 & & \\
    \midrule
    \multirow{9}{*}{\(10^{20}\)}
        & 731M & 172  & \multirow{9}{*}{\(2^{22}\)} & \multirow{9}{*}{\(8.3\times10^{-4}\)} \\*
        & 1B   & 134  & & \\*
        & 1.1B & 122  & & \\*
        & 1.4B & 99.4 & & \\*
        & 1.6B & 82.9 & & \\*
        & 1.8B & 71.0 & & \\*
        & 2.8B & 51.8 & & \\*
        & 3.2B & 44.4 & & \\*
        & 7.5B & 19.4 & & \\
\end{longtable}
\endgroup

\newpage
\raggedbottom

\begin{table}[H]
    \centering
    \small
    \setlength{\tabcolsep}{8pt}
    \renewcommand{\arraystretch}{1.12}
    \caption{\textbf{Common training configurations for the MoE architecture sweeps.}
    At each reference budget \(C\), the allocation targets \(M^*(C)\) and \(D^*(C)\) are obtained from the fitted compute-allocation laws. All candidates are trained for \(3D^*(C)\) nominal tokens, corresponding to approximately \(3C\) training FLOPs, while \(B\) and \(\eta\) are obtained by evaluating the fitted hyperparameter laws at \(3C\).}
    \label{tab:architecture-training-configurations}
    \begin{tabular*}{\textwidth}{@{\extracolsep{\fill}}lcccc@{}}
        \toprule
        \(C\) & \(M^*(C)\) & \(3D^*(C)\) & \(B\) & \(\eta\) \\
        \midrule
        \(6\times10^{17}\) & \(1.43\times10^8\) & \(1.26\times10^{10}\) & \(1048576\) & \(2.2\times10^{-3}\) \\
        \(2\times10^{18}\) & \(2.54\times10^8\) & \(2.36\times10^{10}\) & \(1310720\) & \(1.7\times10^{-3}\) \\
        \(6\times10^{18}\) & \(4.28\times10^8\) & \(4.20\times10^{10}\) & \(2097152\) & \(1.3\times10^{-3}\) \\
        \(2\times10^{19}\) & \(7.59\times10^8\) & \(7.91\times10^{10}\) & \(3145728\) & \(9.4\times10^{-4}\) \\
        \(6\times10^{19}\) & \(1.28\times10^9\) & \(1.41\times10^{11}\) & \(4194304\) & \(7.0\times10^{-4}\) \\
        \bottomrule
    \end{tabular*}
\end{table}

\begin{table}[H]
    \centering
    \small
    \setlength{\tabcolsep}{5pt}
    \renewcommand{\arraystretch}{1.12}
    \caption{\textbf{Configurations for the activation-ratio sweep.}
    The upper block reports the backbone and fixed expert settings at each reference compute budget. The lower block lists the activation-ratio candidates shared across budgets, with each column giving one corresponding \((n_e,A)\) pair.}
    \label{tab:architecture-activation-configurations}
    \begin{tabular*}{\textwidth}{@{\extracolsep{\fill}}lccccccc@{}}
        \toprule
        \multicolumn{8}{c}{Backbone and fixed expert settings} \\
        \cmidrule(lr){1-8}
        Compute
        & \(n_{\mathrm{layer}}\)
        & \(d_{\mathrm{model}}\)
        & \(n_{\mathrm{heads}}\)
        & \(n_{\mathrm{kvheads}}\)
        & \(n_a\)
        & \(n_s\)
        & \(d_{\mathrm{expert}}\) \\
        \midrule
        \(6\times10^{17}\) & 8  & 256 & 8  & 2 & 2 & 1 & 256 \\
        \(2\times10^{18}\) & 8  & 448 & 8  & 2 & 2 & 1 & 448 \\
        \(6\times10^{18}\) & 10 & 512 & 16 & 4 & 2 & 1 & 512 \\
        \(2\times10^{19}\) & 12 & 640 & 16 & 4 & 2 & 1 & 640 \\
        \(6\times10^{19}\) & 15 & 768 & 16 & 4 & 2 & 1 & 768 \\
        \bottomrule
    \end{tabular*}

    \vspace{0.5em}

    \begin{tabular*}{\textwidth}{@{\extracolsep{\fill}}l*{8}{c}@{}}
        \toprule
        \multicolumn{9}{c}{Activation-ratio candidates} \\
        \cmidrule(lr){1-9}
        Candidate   & 1   & 2  & 3    & 4    & 5   & 6   & 7   & 8 \\
        \midrule
        \(n_e\)     & 2   & 4  & 8    & 16   & 32  & 64  & 128 & 256 \\
        \(A\) (\%)  & 100 & 60 & 33.3 & 17.6 & 9.1 & 4.6 & 2.3 & 1.2 \\
        \bottomrule
    \end{tabular*}
\end{table}

\begin{table}[H]
    \centering
    \small
    \setlength{\tabcolsep}{5pt}
    \renewcommand{\arraystretch}{1.12}
    \caption{\textbf{Configurations for the expert-granularity sweep.}
    The upper block reports the Transformer backbone at each reference compute budget. In the lower block, each column gives one candidate's \(G\), \(n_e\), \(n_a\), and \(n_s\), together with the corresponding \(d_{\mathrm{expert}}\) at every budget.}
    \label{tab:architecture-granularity-configurations}
    \begin{tabular*}{\textwidth}{@{\extracolsep{\fill}}lcccc@{}}
        \toprule
        \multicolumn{5}{c}{Transformer backbones} \\
        \cmidrule(lr){1-5}
        Compute
        & \(n_{\mathrm{layer}}\)
        & \(d_{\mathrm{model}}\)
        & \(n_{\mathrm{heads}}\)
        & \(n_{\mathrm{kvheads}}\) \\
        \midrule
        \(6\times10^{17}\) & 8  & 256 & 8  & 2 \\
        \(2\times10^{18}\) & 8  & 448 & 8  & 2 \\
        \(6\times10^{18}\) & 10 & 512 & 16 & 4 \\
        \(2\times10^{19}\) & 12 & 640 & 16 & 4 \\
        \(6\times10^{19}\) & 15 & 768 & 16 & 4 \\
        \bottomrule
    \end{tabular*}

    \vspace{0.5em}

    \begin{tabular*}{\textwidth}{@{\extracolsep{\fill}}l*{7}{c}@{}}
        \toprule
        \multicolumn{8}{c}{Expert-granularity candidates} \\
        \cmidrule(lr){1-8}
        Candidate & 1 & 2 & 3 & 4 & 5 & 6 & 7 \\
        \midrule
        \(G\)   & 2  & 4   & 6   & 8   & 12  & 16  & 20 \\
        \(n_e\) & 64 & 128 & 192 & 256 & 384 & 512 & 640 \\
        \(n_a\) & 2  & 4   & 6   & 8   & 12  & 16  & 20 \\
        \(n_s\) & 1  & 2   & 3   & 4   & 6   & 8   & 10 \\
        \addlinespace[0.3em]
        \(d_{\mathrm{expert}}(6\times10^{17})\) & 256 & 128 & 85  & 64  & 43  & 32 & 26 \\
        \(d_{\mathrm{expert}}(2\times10^{18})\) & 448 & 224 & 149 & 112 & 75  & 56 & 45 \\
        \(d_{\mathrm{expert}}(6\times10^{18})\) & 512 & 256 & 171 & 128 & 85  & 64 & 51 \\
        \(d_{\mathrm{expert}}(2\times10^{19})\) & 640 & 320 & 213 & 160 & 107 & 80 & 64 \\
        \(d_{\mathrm{expert}}(6\times10^{19})\) & 768 & 384 & 256 & 192 & 128 & 96 & 77 \\
        \bottomrule
    \end{tabular*}
\end{table}

\begin{table}[H]
    \centering
    \small
    \setlength{\tabcolsep}{5pt}
    \renewcommand{\arraystretch}{1.12}
    \caption{\textbf{Configurations for the shared-expert-ratio sweep.}
    The upper block reports the backbone and fixed expert settings at each reference compute budget. The lower block lists the shared-expert-ratio candidates shared across budgets, with each column giving one corresponding \((n_a,n_s,S)\) tuple.}
    \label{tab:architecture-shared-configurations}
    \begin{tabular*}{\textwidth}{@{\extracolsep{\fill}}lcccccc@{}}
        \toprule
        \multicolumn{7}{c}{Backbone and fixed expert settings} \\
        \cmidrule(lr){1-7}
        Compute
        & \(n_{\mathrm{layer}}\)
        & \(d_{\mathrm{model}}\)
        & \(n_{\mathrm{heads}}\)
        & \(n_{\mathrm{kvheads}}\)
        & \(n_e\)
        & \(d_{\mathrm{expert}}\) \\
        \midrule
        \(6\times10^{17}\) & 8  & 256 & 8  & 2 & 256 & 64 \\
        \(2\times10^{18}\) & 8  & 448 & 8  & 2 & 256 & 112 \\
        \(6\times10^{18}\) & 10 & 512 & 16 & 4 & 256 & 128 \\
        \(2\times10^{19}\) & 12 & 640 & 16 & 4 & 256 & 160 \\
        \(6\times10^{19}\) & 15 & 768 & 16 & 4 & 256 & 192 \\
        \bottomrule
    \end{tabular*}

    \vspace{0.5em}

    \begin{tabular*}{\textwidth}{@{\extracolsep{\fill}}l*{7}{c}@{}}
        \toprule
        \multicolumn{8}{c}{Shared-expert-ratio candidates} \\
        \cmidrule(lr){1-8}
        Candidate   & 1  & 2   & 3    & 4    & 5  & 6    & 7 \\
        \midrule
        \(n_a\)     & 12 & 11  & 10   & 8    & 6  & 4    & 2 \\
        \(n_s\)     & 0  & 1   & 2    & 4    & 6  & 8    & 10 \\
        \(S\) (\%)  & 0  & 8.3 & 16.7 & 33.3 & 50 & 66.7 & 83.3 \\
        \bottomrule
    \end{tabular*}
\end{table}

\end{document}